\documentclass[twoside]{article}

\PassOptionsToPackage{numbers}{natbib}

\usepackage[preprint]{aistats2026}

\usepackage[round]{natbib}
\renewcommand{\bibname}{References}
\renewcommand{\bibsection}{\subsubsection*{\bibname}}

\usepackage{microtype}
\usepackage{graphicx}
\usepackage{booktabs} %

\usepackage{hyperref}

\usepackage{tikz}
\usepackage{amsmath}
\usepackage{subcaption}
\usepackage{multirow}
\usepackage{svg}
\usepackage{hyperref}
\usepackage{url}
\usepackage{booktabs}
\usepackage{threeparttable}
\usepackage{wrapfig}
\usepackage{graphicx}
\usepackage{tabularx}
\usepackage[table]{xcolor} %
\usepackage{dblfloatfix}  %

\usepackage{amsmath}
\usepackage{amssymb}
\usepackage{mathtools}
\usepackage{amsthm}

\usepackage[capitalize,noabbrev]{cleveref}

\usepackage[textsize=tiny]{todonotes}

\usepackage{amsmath,amsfonts,bm}

\def\eqref#1{equation~\ref{#1}}

\def\1{\bm{1}}

\DeclareMathAlphabet{\mathsfit}{\encodingdefault}{\sfdefault}{m}{sl}
\SetMathAlphabet{\mathsfit}{bold}{\encodingdefault}{\sfdefault}{bx}{n}

\newcommand{\ie}{\textit{i.e.}}

\definecolor{RowGray}{RGB}{220,220,220} %
\definecolor{RowTextGray}{gray}{0.45} %

\newif\ifshowcomments

\ifshowcomments
    \newcommand{\waleed}[1]{{\color{red}[Waleed: #1]}}
    \newcommand{\krishna}[1]{{\color{blue}[Krishna: #1]}}
\else
    \newcommand{\waleed}[1]{}
    \newcommand{\krishna}[1]{}
\fi

\newcommand{\squishlist}{
 \begin{list}{$\bullet$}
 		{ \setlength{\itemsep}{0pt}
 			\setlength{\parsep}{3pt}
 			\setlength{\topsep}{3pt}
 			\setlength{\partopsep}{0pt} 
 			\setlength{\leftmargin}{1.5em}
 			\setlength{\labelwidth}{1em}
 			\setlength{\labelsep}{0.5em} } }
\newcommand{\squishend}{
  \end{list}  }

\begin{document}

\twocolumn[

\aistatstitle{How Many Parameters Does Your Task Really Need? Task Specific Pruning with LLM-Sieve }

\aistatsauthor{ Waleed Reda \And Abhinav Jangda \And  Krishna Chintalapudi }

\aistatsaddress{ Microsoft \And  Microsoft \And Microsoft } ]

\begin{abstract}
As Large Language Models (LLMs) are increasingly deployed for narrow tasks in resource-constrained settings, a central question arises: \emph{how much of an LLM is truly necessary for a given task?} 
We present \textsc{LLM-Sieve}, a framework that prunes LLMs down to the minimal parameter subset needed to preserve task performance. 
Our approach introduces two innovations: (i) \emph{output-aligned non-orthogonal projections}, which yield more faithful low-rank approximations than traditional PCA/SVD by aligning directly with layer outputs; and (ii) \emph{adaptive pruning via a Genetic Algorithm}, which automatically discovers matrix-specific pruning levels and exposes the uneven distribution of task-relevant knowledge. 
Across models from 3.8B to 70B parameters, LLM-Sieve removes 20–75\% of weights with only 1–5\% accuracy loss—substantially ahead of prior pruning methods. 
Beyond efficiency, our framework reveals \emph{bottleneck matrices} that concentrate critical knowledge, suggesting architectural implications for future LLM design. 
LLM-Sieve integrates seamlessly with LoRA fine-tuning and quantization, enabling both efficient deployment and deeper understanding of knowledge organization in LLMs.
\end{abstract}

\section{Introduction}
\label{sec:intro}

Large Language Models (LLMs) are trained as billion-parameter generalists, yet in practice they are typically deployed in \emph{narrow} domains such as retrieval-augmented generation (RAG) (e.g., medical or legal QA), classification (e.g., sentiment analysis), or semantic parsing (e.g., entity extraction). This mismatch raises a fundamental question:
\\ \\
\begin{quote}
\textbf{Task-aware parameter sufficiency:} \\
\emph{What is the smallest subset of LLM parameters needed to preserve end-to-end accuracy and reasoning within a fixed task domain?}
\end{quote}

To the best of our knowledge, no prior work has directly addressed this question.
We take the first step with \textbf{LLM-Sieve}, a framework that prunes an LLM to
a compact, task-specialized subnetwork while maintaining accuracy within a
user-specified tolerance~$\epsilon$ (formalized in Section~\ref{sec:design}).

\begin{figure}[ht]
    \vspace{-0.1in}
    \centering
    \includegraphics[width=0.8\linewidth]{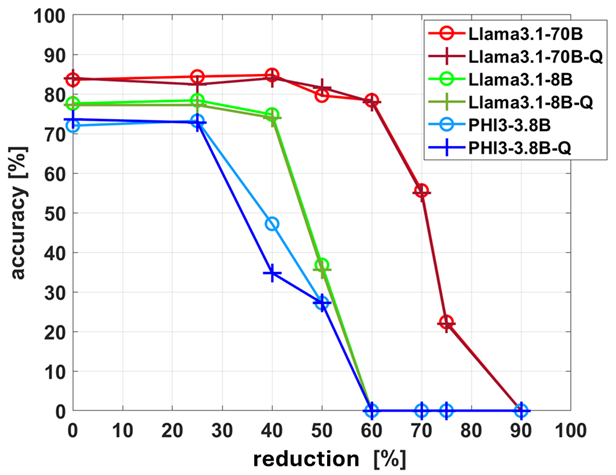}
    \caption{\small End-to-end accuracy vs.\ percentage of parameters pruned for a Generic-RAG multi-step QA task on Phi-3-mini (3.8B), LLaMA-3.1-8B, and LLaMA-3.1-70B, with and without 8-bit quantization (``-Q''). Accuracy remains stable until 25--60\% pruning, then drops sharply, revealing redundancy followed by task-critical bottlenecks.}
    \label{fig:teaserfigure}
    \vspace{-0.08in}
\end{figure}

As shown in Fig.~\ref{fig:teaserfigure}, pruning with LLM-Sieve reveals a two-phase curve: accuracy changes by less than $5\%$ until 25--60\% of parameters are pruned (depending on model), then collapses once bottlenecks are reached. On narrow tasks such as sentiment classification, up to 75\% of parameters can be removed on LLaMA-3.1-70B within a $\leq 5\%$ accuracy gap (Section~\ref{sec:eval-downsize}, Figure~\ref{fig:main_results}).  
Thus, by embracing a task-specific pruning framework, LLM-Sieve far outperforms the state-of-the-art (Section~\ref{sec:eval-comparison}): it achieves 25--75\% pruning under comparable accuracy constraints, whereas prior methods succeed only at 1--10\%.  

LLM-Sieve advances state of the art with two key ideas.

\noindent
First, \textbf{output-aligned non-orthogonal projections}: unlike prior low-rank methods that separately project weights or inputs (e.g., PCA, SVD) and assume aligned subspaces, LLM-Sieve introduces a joint projection aligned with layer outputs. This captures task-specific structure more faithfully and goes beyond the orthogonality constraints of SVD by learning non-orthogonal subspaces tailored to each compression level.  

\noindent
Second, \textbf{adaptive pruning via genetic search}: rather than applying fixed, uniform pruning ratios, LLM-Sieve uses a genetic algorithm to adapt pruning levels across matrices. This both avoids critical bottlenecks and directly optimizes for non-differentiable end-to-end metrics (e.g., GPT-as-a-judge accuracy), enabling compression gains unattainable with gradient-based approaches alone.  
Together, these choices yield substantial compression gains and near-linear latency reductions with minimal accuracy loss.

\textbf{Beyond efficiency, LLM-Sieve provides scientific insights.} It reveals that task-relevant knowledge is unevenly distributed across layers and matrices, with certain components acting as bottlenecks. This suggests that pruning with LLM-Sieve can serve as a probe into how knowledge is organized inside an LLM.  

\textbf{Compatibility with quantization and fine-tuning.} In practical pipelines, quantization is often used to reduce memory and latency, while LoRA is used for fine-tuning. Pruning with LLM-Sieve is complementary—not competitive—to both, enabling a unified pipeline for building compact, task-specialized LLMs (Section~\ref{sec:eval-LoRA}).  

\paragraph{Contributions.}
\begin{itemize}
\setlength{\itemsep}{0.15\baselineskip} %
\item \textbf{Problem framing.} We formalize \emph{task-aware parameter sufficiency}—finding the minimal subnetwork needed for a task under tolerance~$\epsilon$.  
\item \textbf{Method.} A new combination of output-aligned, non-orthogonal projections and GA-based adaptive pruning.  
\item \textbf{Insights.} Empirical discovery of \emph{bottleneck matrices} and highly uneven knowledge concentration across layers/matrix types.  
\item \textbf{Performance.} LLM-Sieve achieves 25--75\% parameter reduction (vs.\ 1--10\% for prior methods), and up to $\sim$90\% effective memory savings when combined with 8-bit quantization, with minimal loss in end-to-end accuracy.  
\end{itemize}

\section{Related Work}
\label{sec:relatedwork}

Research on compressing large language models (LLMs) has explored pruning, low-rank approximation, quantization, and knowledge distillation to reduce size and computation.  

\vspace{-0.1in}
\paragraph{Pruning.}  
Pruning methods can be unstructured or structured. Unstructured pruning removes weights by magnitude~\cite{zafrir2021prune,frankle2018lottery}, yielding sparse matrices but limited GPU speedup due to irregular memory access~\cite{sparsegpt}. SparseGPT~\cite{sparsegpt} and Wanda~\cite{wanda} apply lightweight tuning to recover accuracy after aggressive pruning. Structured pruning eliminates entire channels, heads, or blocks, producing dense submatrices that map better to hardware. LLM-Pruner~\cite{llmpruner} removes less critical structures using gradient saliency. LLM-Surgeon~\cite{llmsurgeon} and Eigendamage~\cite{eigendamage} approximate loss with second-order statistics, but curvature estimation across millions of activations is expensive and hard to scale.  

\vspace{-0.1in}
\paragraph{Low-rank Approximation.}  
Low-rank methods are a well-known variant of structured pruning, and compress by reducing matrix multiplication dimensions~\cite{noach2020compressing}. SliceGPT~\cite{slicegpt} projects input activations into lower-dimensional subspaces, pruning rows or columns. ESPACE~\cite{espace} also uses input-driven low-rank projections but decides how to perform SVD for each matrix using different fidelity metrics. LASER~\cite{laser} decomposes weights directly, while SVD-LLM, ASVD, and M-PIFA~\cite{svdllm,asvd,mpifa} reconstruct from outputs, leveraging other SVD variations. LLM-Sieve builds on this but departs by jointly projecting inputs and outputs via gradient descent, learning non-orthonormal bases optimized for task accuracy. It further introduces adaptive pruning with genetic search to identify task-critical (\emph{bottleneck}) matrices. Table~\ref{tab:compression_methods} compares these approaches.

\vspace{-0.1in}
\paragraph{Other Downsizing Techniques.}  
Quantization reduces precision (e.g., FP32 $\rightarrow$ INT8/FP4)~\cite{gptq,smoothquant,quarot,spinquant}, lowering memory and compute with modest accuracy loss~\cite{gholami2022survey,hoefler2021sparsity}. Knowledge distillation~\cite{hinton2015distilling} transfers knowledge from a larger teacher to a smaller student model. Both are complementary to pruning and can be combined.  

\begin{table*}[ht]
\centering
\small
\begin{tabular}{|l|l|l|}
\hline
\textbf{Method} & \textbf{Low-Rank Method} & \textbf{Rank Selection Strategy} \\ \hline
\rowcolor{RowGray}
\textbf{LLM-Sieve (Ours)} & \textbf{Joint input+weight ($X,W$), non-orthogonal}& \textbf{Adaptive (end-to-end, GA)} \\ 
\rowcolor{RowGray}
& \textbf{minimize $|Y-\tilde{Y}|$ via GD} & \\ \hline
LASER & Weights ($W$) via SVD & Uniform (fixed) \\ \hline
SliceGPT & Inputs ($X$) via PCA & Uniform (fixed) \\ \hline
ESPACE & Inputs ($X$) via PCA (variance-preserving) & Uniform (fixed) \\ \hline
SVD-LLM & Weights ($W$) via SVD (whitening) & Uniform (fixed) \\ \hline
ASVD & Weights ($W$) via SVD (scaling) & Adaptive (perplexity) \\ \hline
M-PIFA & Weights ($W$) via SVD (pivoting) & Adaptive (heuristic) \\ \hline
\end{tabular}
\caption{Comparison of low-rank pruning methods. LLM-Sieve uniquely uses \textbf{end-to-end adaptive} rank selection via GA and learns \textbf{output-aligned} projections.}
\label{tab:compression_methods}
\end{table*}

\section{Transformer Architecture Background}
\label{sec:background}
\begin{figure}[t]
    \centering
    \includegraphics[width=\linewidth]{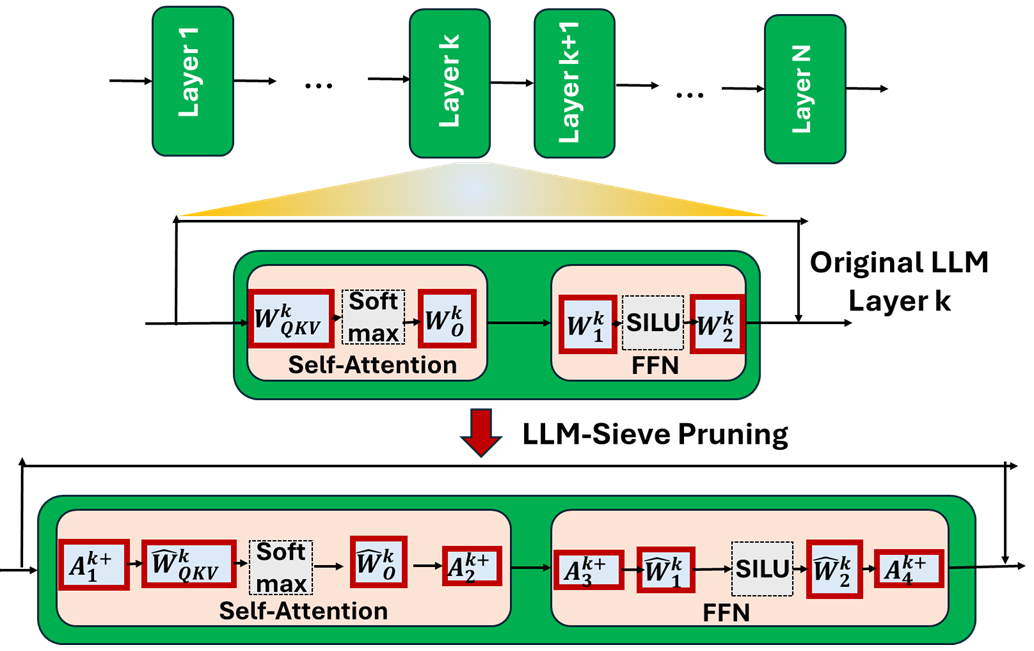}
    \caption{\small Each matrix multiplication in an LLM is approximated in LLM-Sieve.}
    \label{fig:LLM-Sieve-HighLevel}
\end{figure}

In this section, we provide the necessary background on transformer networks~\cite{attention}, required for the rest of the paper. The transformer architecture (Figure~\ref{fig:LLM-Sieve-HighLevel}) comprises  a series of layers, each composed of a multi-head self-attention mechanism followed by a feed-forward network (FFN) block. Between these blocks, normalization layers such as LayerNorm \cite{layernorm} or RMSNorm \cite{rmsnorm} are applied, often in conjunction with residual connections.

Inference begins with an embedding layer that transforms input token IDs and position IDs into dense vectors.  After embeddings, the signal matrix $X\in\Re^{N\times D}$, where $D$ denotes the embedding dimension and $N$ the sequence length, is passed through a LayerNorm operation which normalizes each row of $X$. After this $X$ is transformed as it passes through in each of the layers. 

In the attention block at the $k^{th}$ layer, the signal is projected into key ($K^k = XW^k_K$), query ($Q^k = XW^k_Q$), and value ($V^k = XW^k_V$) matrices using respective weight matrices $W^k_K$, $W^k_Q$, and $W^k_V$, usually concatenated into a single matrix $W^k_{QKV}$ for efficient computation. The self-attention mechanism computes the attention scores:
\begin{equation}
\small{
\text{Attention}(Q^k, K^k, V^k) = \text{softmax}\left(\frac{Q^k(K^k)^T}{\sqrt{D}}\right)V^k
}
\end{equation}
Multi-head attention extends this by performing these operations in parallel across $h$ heads, concatenating their outputs, and applying a final projection:
\begin{equation}
\small{
\text{MultiHead}(Q^k, K^k, V^k) = \text{Concat}(\text{head}_1, \ldots, \text{head}_h)W^k_O
}
\end{equation}
where $W^k_O$ is called the output projection matrix.

The FFN block typically consists of two linear transformations separated by a non-linear activation function, such as ReLU or GeLU. The operation of an FFN block is given by:
\begin{equation}
\text{FFN}(X) = \sigma(XW^k_1 + b^k_1)W^k_2 + b^k_2
\end{equation}
where $W^k_1$ and $W^k_2$ are weight matrices, $b^k_1$ and $b^k_2$ are biases, and $\sigma$ represents the activation function.

These blocks are repeated in every layer, and the embedding output has to pass through all the layers.  Due to the autoregressive nature of transformers, outputs are fed back and multiple forward passes are required to produce the output to the prompt.

\section{LLM-Sieve}
\label{sec:design}
\emph{LLM-Sieve aims to prune a Large Language Model (LLM) while ensuring task performance degrades by at most a user-specified tolerance $\epsilon$.} Let $L(\Theta)$ denote the original LLM with parameters $\Theta$, mapping inputs $\mathbf{X}$ to outputs $\mathbf{Y} = L(\mathbf{X} \mid \Theta)$. A task-specific evaluation function $e(\mathbf{Y})$ quantifies quality—e.g., F1 score or a learned metric such as GPT-4o-as-a-judge. The objective is to construct a pruned model $\hat{L}(\hat{\Theta})$ such that $e\!\left(\hat{L}(\mathbf{X}_{\mathcal{C}} \mid \hat{\Theta})\right) \ge a_0$ on a calibration set $\mathbf{X}_{\mathcal{C}}$, where $a^* = e\!\left(L(\mathbf{X}_{\mathcal{C}} \mid \Theta)\right)$ is the original performance and $a_0$ is chosen so that $\frac{a^* - a_0}{a^*} = \epsilon$. Equivalently, the pruned model preserves at least $(1-\epsilon)$ of the original task-specific performance.

\begin{figure}[t]
    \centering
    \includegraphics[width=\linewidth]{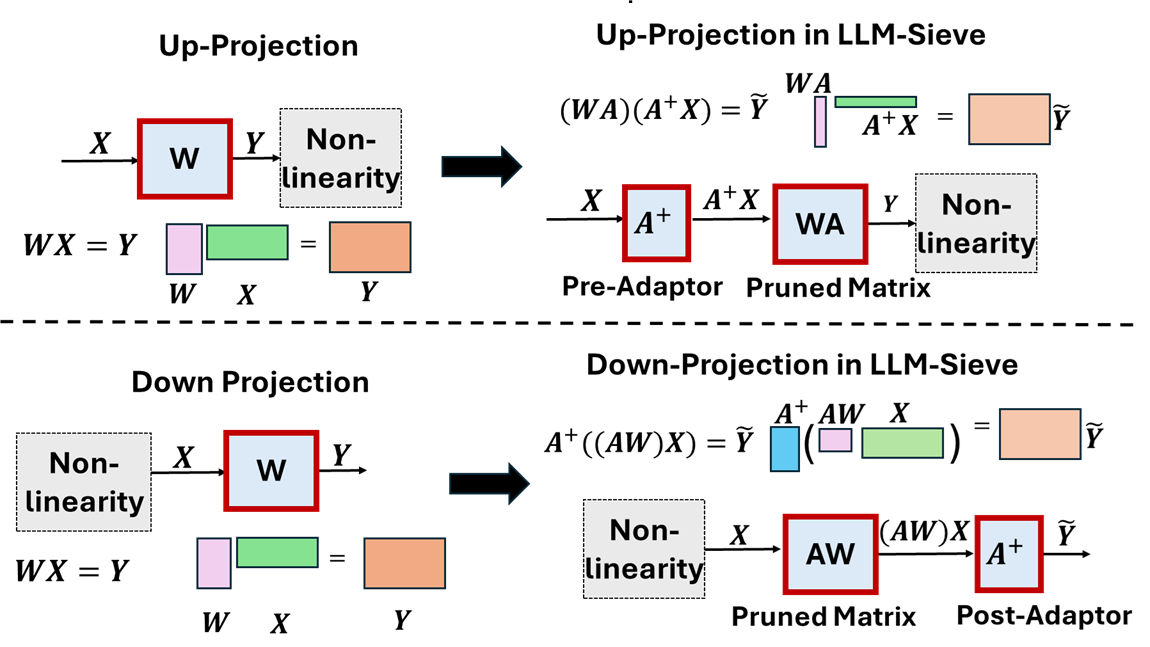}
    \caption{\small Low-rank approximations used in LLM-Sieve pruning.}
    \label{fig:approximateMultiplication}
\end{figure}

\subsection{LLM-Sieve Pruning}
\label{sec:LLM-Sieve-Prune}

LLM-Sieve approximates each matrix multiplication in a task-specific low-rank subspace (Figure~\ref{fig:LLM-Sieve-HighLevel}). Because surrounding layers either expand or reduce dimensionality, we distinguish between up-projection and down-projection multiplications.

\noindent
\textbf{Up-projection multiplications (Figure \ref{fig:approximateMultiplication}).}  
For matrices that expand dimensionality (e.g., $\mathbf{W}_{QKV}$ in attention or $\mathbf{W}_1$ in FFN), the adapter must be applied \emph{before} the projection, while $\mathbf{X}$ is still low-dimensional. We approximate
\[
\tilde{\mathbf{Y}} = (\mathbf{W}\mathbf{A})(\mathbf{A}^\dagger \mathbf{X}),
\]
where $\mathbf{A}\in \mathbb{R}^{R \times H}$ with $R<D$. At inference, inputs are first projected down $\hat{\mathbf{X}}=\mathbf{A}^\dagger \mathbf{X}\in \mathbb{R}^R$, and outputs are computed as $\tilde{\mathbf{Y}}=\hat{\mathbf{W}}\hat{\mathbf{X}}$ with $\hat{\mathbf{W}}=\mathbf{W}\mathbf{A}$.

\noindent
\textbf{Down-projection multiplications (Figure \ref{fig:approximateMultiplication}).}  
For matrices that reduce dimensionality (e.g., $\mathbf{W}_2$ in FFN), the adapter is applied \emph{after} the projection, when outputs are already low-dimensional:
\[
\tilde{\mathbf{Y}} = \mathbf{A}^\dagger\big((\mathbf{A}\mathbf{W})\mathbf{X}\big),
\]
with $\mathbf{A}\in \mathbb{R}^{R\times D}$. Here computations are performed in $\mathbb{R}^R$ before projecting back up.

\begin{figure*}[t]
    \centering
    \begin{minipage}[t]{0.26\textwidth}
        \centering
        \includegraphics[width=\linewidth]{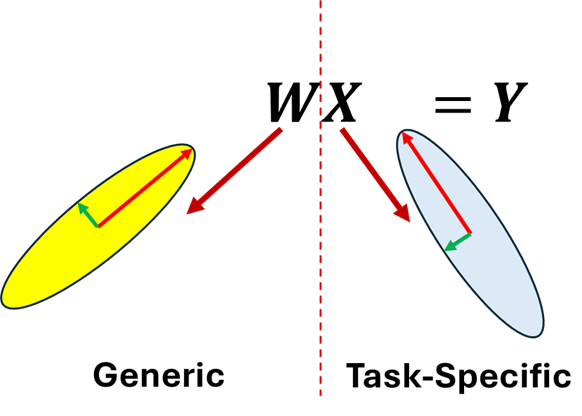}
        \captionof{figure}{\small Intuition behind LLM-Sieve projections.}
        \label{fig:LLM-Sieve-Intuition}
    \end{minipage}
    \hfill
    \begin{minipage}[t]{0.70\textwidth}
        \centering
        \includegraphics[width=\linewidth]{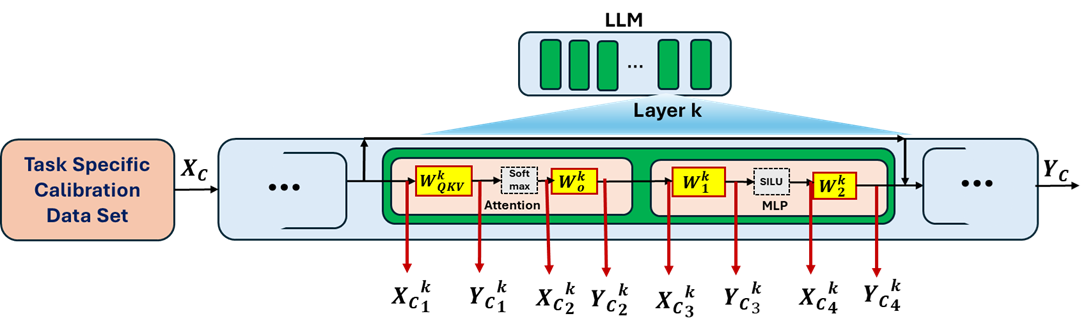}
        \captionof{figure}{\small Calibration step in LLM-Sieve.}
        \label{fig:calibrationStep}
    \end{minipage}
\end{figure*}

\noindent
\textbf{Intuition.}
Standard methods reduce only one side of the multiplication: SVD compresses $\mathbf{W}$, while PCA compresses inputs $\mathbf{X}$. Both implicitly assume that the chosen subspace will align with the other, an assumption that rarely holds in practice since $\mathbf{W}$ is task-agnostic and $\mathbf{X}$ is task-specific (Figure~\ref{fig:LLM-Sieve-Intuition}). As a result, these approaches are limited. In contrast, LLM-Sieve learns task-aware projections that directly approximate the outputs $\mathbf{Y} = \mathbf{W}\mathbf{X}$ by minimizing reconstruction error $\|\mathbf{Y} - \tilde{\mathbf{Y}}\|$.

\noindent
\textbf{Calibration.}  
Adaptor matrices $\mathbf{A}_i^k$ are learned using a calibration dataset $\mathbf{X}_{\mathcal{C}}$, by minimizing reconstruction error between true outputs $\mathbf{Y}_{\mathcal{C}}^k$ and approximations $\tilde{\mathbf{Y}}_{\mathcal{C}}^k$ (Figure \ref{fig:calibrationStep}). For example, LLaMA-3.1-70B has 80 layers $\times$ 4 multiplications each ($\mathbf{W}_{QKV}, \mathbf{W}_o, \mathbf{W}_1, \mathbf{W}_2$), yielding 320 adaptor matrices. Each adaptor matrix is trained for 2 epochs over a 200K-token calibration set, with a batch size of 5K, using the Adam optimizer with a learning rate of 0.001. At each step, we retain the adaptor matrices that minimize the L2 reconstruction error. Early stopping is not used.

\noindent
\textbf{Pruning factor.}  
For $\mathbf{W}\in\mathbb{R}^{H\times D}$ compressed to rank $R$, the pruned form stores $\hat{\mathbf{W}}\in\mathbb{R}^{R\times H}$ and $\mathbf{A}\in\mathbb{R}^{R\times D}$, totaling $R(H+D)$ parameters. The pruning factor is therefore
\[
p = \frac{R(H+D)}{DH}.
\]

\subsection{Adaptive Pruning}
\label{design:GA}

LLM-Sieve pruning (Section~\ref{sec:LLM-Sieve-Prune}) requires specifying the intermediate rank $R$ for each adaptor matrix $\mathbf{A}$. In an $N$-layer model, there are $4N$ matrix multiplications ($\mathbf{W}_{QKV}, \mathbf{W}_o, \mathbf{W}_1, \mathbf{W}_2$ per layer), and their sensitivity to pruning varies. To capture this, we define a pruning factor vector $\mathbf{p} = \langle p_1^1, \ldots, p_4^N \rangle \in [0,1]^{4N}$, where $p_i^k$ is the fraction of parameters retained in matrix $i$ of layer $k$ of the LLM.

The objective is to minimize parameter count subject to an end-to-end performance tolerance $\epsilon$:
\begin{equation}
\min_{\mathbf{p}} \;\; \|\hat{\Theta}(\mathbf{p})\|_0 
\quad \text{s.t.} \quad
\frac{a^* - e(\hat{L}(\mathbf{X}_\mathcal{C} \mid \hat{\Theta}(\mathbf{p})))}{a^*} \leq \epsilon,
\label{eq:llm-sieve-objective}
\end{equation}
where $a^*$ is the unpruned baseline and $e(\cdot)$ may be non-differentiable (e.g., GPT-4o-as-judge).

\subsubsection{Uniform Pruning (Baseline)}
\label{sec:URP}
A common baseline is to prune all matrices by the same factor $p^*$. In this setup, LLM-Sieve automates the choice of $p^*$ via binary search: starting with bounds $(p_{\text{low}},p_{\text{up}})$, the midpoint $p_{\text{mid}}$ is evaluated for our task. If performance meets the tolerance, $p_{\text{up}}\!\leftarrow\!p_{\text{mid}}$; otherwise $p_{\text{low}}\!\leftarrow\!p_{\text{mid}}$. This repeats until convergence.

\subsubsection{Adaptive Pruning with GA}
\label{sec:genetic}
Unlike uniform pruning, adaptive pruning assigns each matrix its own pruning ratio. Optimizing the pruning factor vector $\mathbf{p}$ is a high-dimensional, non-differentiable problem, which makes gradient-based or exhaustive search infeasible. We therefore employ a Genetic Algorithm (GA), an evolutionary search method well-suited for discrete, combinatorial spaces. At a high level, GA maintains a population of candidate pruning vectors, evaluates their task performance, and iteratively improves them through selection, crossover, and mutation. Over successive generations, the population converges toward pruning configurations that maximize compression while respecting the accuracy tolerance.

\noindent
\textbf{Chromosome Encoding.}  
Each chromosome encodes a pruning vector $\mathbf{p}$. A population of $M{=}100$ chromosomes is initialized with factors randomly sampled from $\{1,0.9,0.75,0.6,0.5,0.35,0.25,0.2,0.1,0.05\}$; 10 chromosomes are initialized with uniform pruning.

\noindent
\textbf{Crossover.}  
Parent chromosomes are selected proportional to fitness, split at a random point, and recombined. Crossover probability is 0.5.  

\noindent
\textbf{Mutation.}  
With probability 0.2, a pruning factor is perturbed by one step in the predefined set.  

\noindent
\textbf{Fitness.} 
We use a fitness function with a thresholded exponential penalty:
\[
F(\mathbf{p},a) = c(\mathbf{p}) \times \big(1 + e^{50(a-a_0)}\big),
\]
where $c(\mathbf{p})$ is the overall compression ratio and $a$ the measured task accuracy. Configurations above threshold $a_0$ are rewarded, while underperforming ones are penalized exponentially. The GA terminates when no $\ge$5\% improvement is observed for 10 generations.

\noindent

\textbf{Parallelism.}  
Since pruning operations are independent, pruned matrices are precomputed in parallel, and chromosome evaluations within each GA generation are also parallelized, making GA search highly scalable.

\section{Results and Insights}
\label{sec:evaluation}

We apply \textbf{LLM-Sieve} to three models of increasing scale—Phi-3-mini (3.8B), LLaMA-3.1 (8B), and LLaMA-3.1 (70B)—across three representative tasks:  
{\bf (i) Generic RAG}, {\bf (ii) Medical RAG}, and {\bf (iii) Sentiment Analysis}.  
The first two involve answering questions given retrieved context passages, while the last is a binary classification. Together, these tasks span a spectrum of reasoning and output complexity.

\vspace{-0.1in}
\paragraph{Datasets.}
Standard benchmarks suites such as GLUE~\cite{glue}, HELM~\cite{helm}, LongBench~\cite{longbench}, BLURB~\cite{blurb}, MultiMedQA~\cite{multimedqa}, and KILT~\cite{kilt} span many domains, making them unsuitable for evaluating \emph{task-specific sufficiency}. Since our goal is to study redundancy and bottlenecks within narrow domains, we instead draw focused subsets from these suites. For Generic RAG we use \emph{HotpotQA}~\cite{hotpotqa} (Gen RAG-I) and \emph{NaturalQA}~\cite{naturalqa} (Gen RAG-II). 
For Medical RAG we use \emph{PubMedQA}~\cite{pubmedqa} (Med RAG-I) and \emph{MedMCQA}~\cite{medmcqa} (Med RAG-II). 
For Sentiment Analysis we use \emph{IMDB}~\cite{imdb} (Sentiment-I) and \emph{SST2}~\cite{sst2} (Sentiment-II). 
These targeted datasets allow us to probe pruning behavior and task sufficiency in more controlled, domain-specific settings.

\begin{figure*}[t]
    \centering
    \begin{minipage}[t]{1.32\columnwidth}
        \centering
        \includegraphics[width=\linewidth]{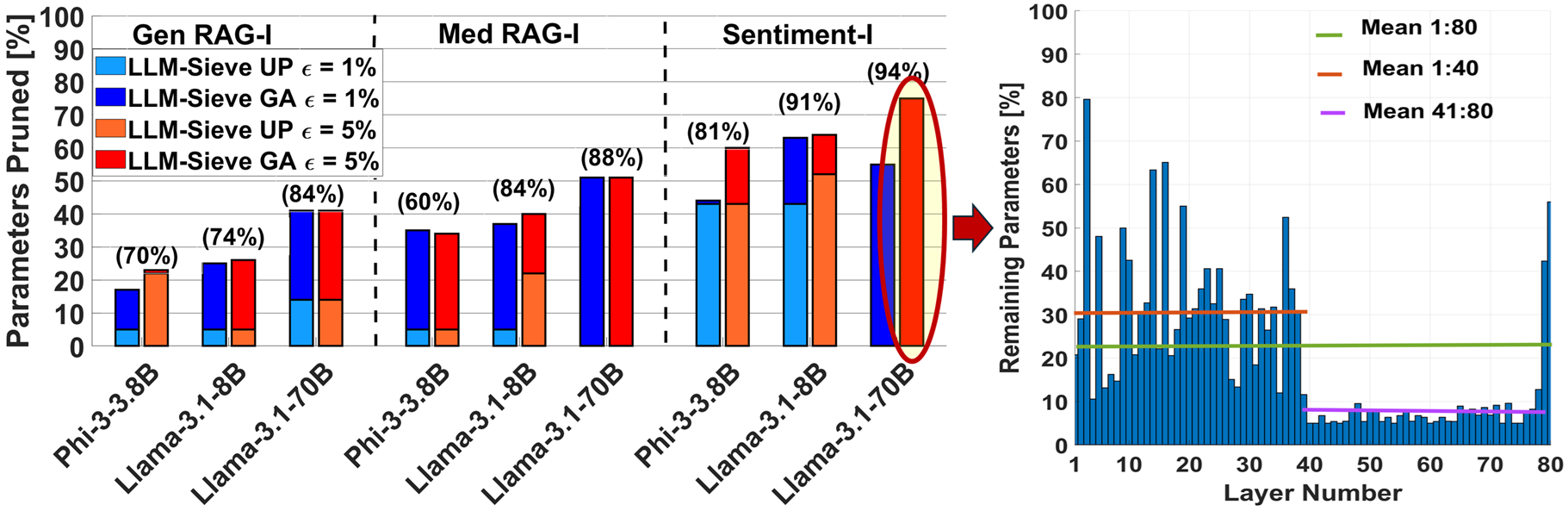}
        \caption{\small{Left: Parameter reduction achieved by LLM-Sieve with and without GA across tasks and models; accuracy is provided in (.) on the top. Right: layer-wise retention for LLaMA-3.1-70B (Sentiment-I), showing that later layers are highly redundant for the task.}}
        \label{fig:main_results}
    \end{minipage}
    \hfill
    \begin{minipage}[t]{0.64\columnwidth}
        \centering        
        \includegraphics[width=\linewidth]{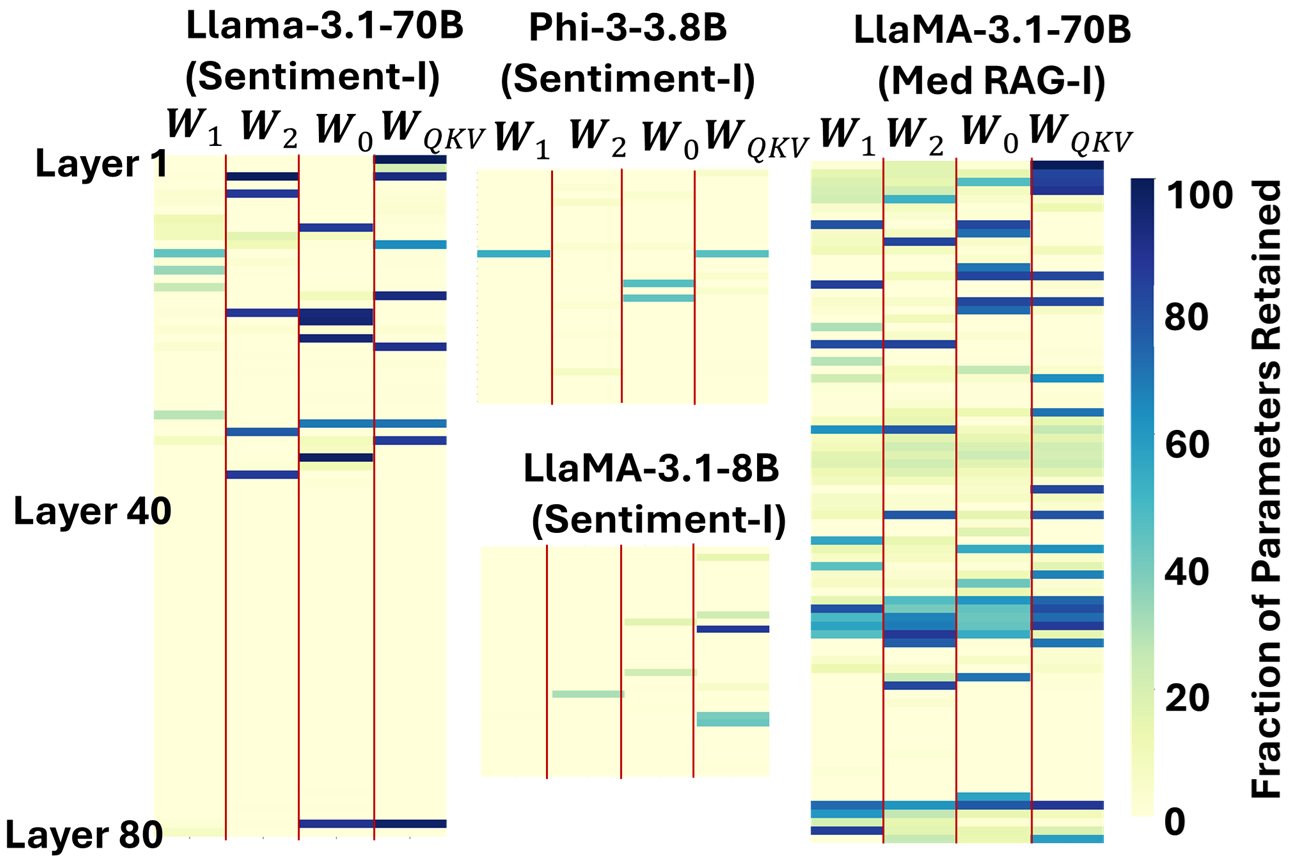}
        \caption{\small{Matrix-wise retention. Dark blue = bottlenecks resistant to pruning.}}
        \label{fig:LLM-MRI}
    \end{minipage}
    \vspace{-0.15in}
\end{figure*}

\vspace{-0.1in}
\paragraph{Measuring Accuracy (Strict, End-to-End).}
We evaluate one-shot accuracy using GPT-4o-as-a-judge~\cite{judge}, which better captures task success than token-level overlap or likelihood metrics such as Exact Match, F1~\cite{f1score}, or perplexity. Unlike these metrics, the judge can recognize semantic equivalence despite paraphrasing, verify that reasoning is factually correct, and require grounding to the provided context.

For each prompt, GPT-4o compares the model’s response to the ground truth under a fixed rubric and returns a binary label (Correct/Incorrect). Accuracy is then:
\[
\mathrm{Accuracy}=\frac{\#\text{ of Correct Responses}}{\#\text{ of Prompts}}\,
\]

For \emph{Generic-RAG} and \emph{Medical-RAG}, a response is marked \emph{Correct} only if \emph{both} (i) the final answer matches the ground truth \emph{and} (ii) the reasoning/justification is judged factually correct and grounded in the supplied passages—this is a stricter end-to-end criterion than the answer-only F1 or perplexity commonly used in pruning evaluations. For \emph{Sentiment Analysis}, the predicted class label must match the ground truth. We provide the judge's system prompts and rubric in Appendix~\ref{appendix:prompts} and set the judge temperature to 0; ambiguous cases are treated as \emph{Incorrect}.

\vspace{-0.1in}
\paragraph{Calibration Data Sufficiency.}
We conduct a calibration-size sensitivity analysis by systematically increasing the number of calibration tokens and monitoring all primary metrics; beyond $200{,}000$ tokens, results stabilize (no material change). Details and curves are provided in Appendix~\ref{appendix:sensitivity}. Accordingly, our calibration sets use $200{,}000$ tokens drawn from \emph{Gen RAG-I}, \emph{Med RAG-I}, and \emph{Sentiment-I}.

\vspace{-0.1in}
\paragraph{Generalization to Unseen Datasets.}
To evaluate out-of-distribution generalization, we reserve disjoint datasets—\emph{Gen RAG-II}, \emph{Med RAG-II}, and \emph{Sentiment-II}—that share no prompts with the calibration sets. We report the same metrics under the strict judge protocol, with no further tuning or recalibration on these held-out datasets.

\vspace{-0.1in}
\paragraph{Baselines.}
We compare two versions of our method—\textbf{LLM-Sieve-GA} (selective pruning with a Genetic Algorithm) and \textbf{LLM-Sieve-UP} (uniform pruning with binary search)—against seven state-of-the-art pruning techniques. Weight-based SVD approaches include \textbf{LASER}~\cite{laser}, \textbf{SVD-LLM}~\cite{svdllm}, \textbf{ASVD}~\cite{asvd}, and \textbf{M-PIFA}~\cite{mpifa}. Input-based approaches include \textbf{SliceGPT}~\cite{slicegpt} and \textbf{ESPACE}~\cite{espace}.
\textbf{LLM-Pruner}~\cite{llmpruner} is a popular gradient-based structured pruning technique that does not use low-rank methods. 

\vspace{-0.1in}
\paragraph{Setup.}
We implement LLM-Sieve on VLLM~\cite{vllm} for memory-efficient inference. Pruning was performed on 8 H100 GPUs in a bare-metal DGX system with pipeline (but not tensor) parallelism to capture per-matrix activations, while GA experiments were run on 96 A100 GPUs across 12 VMs.

\subsection{How Many Parameters Can Be Removed?}
\label{sec:eval-downsize}
As shown in Figure~\ref{fig:main_results}, the removable parameter count depends on the original model size and the narrowness of the task: larger models and narrower tasks permit greater reduction. For narrow classification tasks (e.g., \emph{Sentiment}), which do not require long-form generation, \textsc{LLM-Sieve} on \textsc{LLaMA-3.1-70B} removes \(\mathbf{55\%}\) and \(\mathbf{75\%}\) of parameters while keeping accuracy within \(\epsilon\!\le\!1\%\) and \(\epsilon\!\le\!5\%\) of the baseline, respectively (where \(\epsilon\) denotes absolute accuracy gap). \emph{Even for long-form tasks such as \emph{Medical-RAG} and \emph{Generic-RAG}, the method removes \(\mathbf{20\%\!-\!50\%}\) of parameters while maintaining strong accuracy, suggesting that a substantial fraction of memorized factual detail may be redundant for these tasks.}

\subsection{Adaptive Pruning as a Probe into Uneven Knowledge Distribution}
Uniform, fixed-rank pruning achieves limited safe reduction (often \(<\!5\%\) under fixed-accuracy constraints). Most gains come from \emph{selective} pruning: \textsc{LLM-Sieve-GA} consistently contributes an additional \(\mathbf{10\%\!-\!50\%}\) reduction by adapting to the uneven distribution of task-relevant knowledge across layers and matrices.  

\textbf{Layer-wise specialization.} On \textsc{LLaMA-3.1-70B} for \emph{Sentiment-I} (Fig. \ref{fig:main_results}, right), layers 41--79 retain under 10\% of parameters, layers 1--40 retain at least 30\%, and the first/last layers preserve 70--80\%. This suggests functional specialization: early layers support parsing and reasoning, final few layers drive language realization, while many later layers—important for long-form generation—can be significantly pruned in classification tasks.

\textbf{Matrix-level bottlenecks.} Figure~\ref{fig:LLM-MRI} shows that certain matrices—especially \(\mathbf{W}_{QKV}\) in early and late attention blocks—resist pruning, acting as \emph{bottlenecks} that concentrate task-critical signal. Others, such as the first feed-forward projection \(W_1\), are far more compressible (Appendix~\ref{appendix:bottleneck}).  

Together, these findings explain why uniform pruning underperforms: it ignores where knowledge is dense versus redundant. Beyond efficiency, pruning also serves as a scientific probe into LLM structure, revealing uneven knowledge concentration across layers and matrices. 

These observations further suggest {\bf architectural design implications}: future models could allocate capacity non-uniformly—placing more in bottleneck components and less in redundant ones—reducing redundancy already at training time rather than only after pruning.

\begin{figure*}[t]
    \centering
    \begin{minipage}[t]{0.64\columnwidth}
        \centering
        \includegraphics[width=\linewidth]{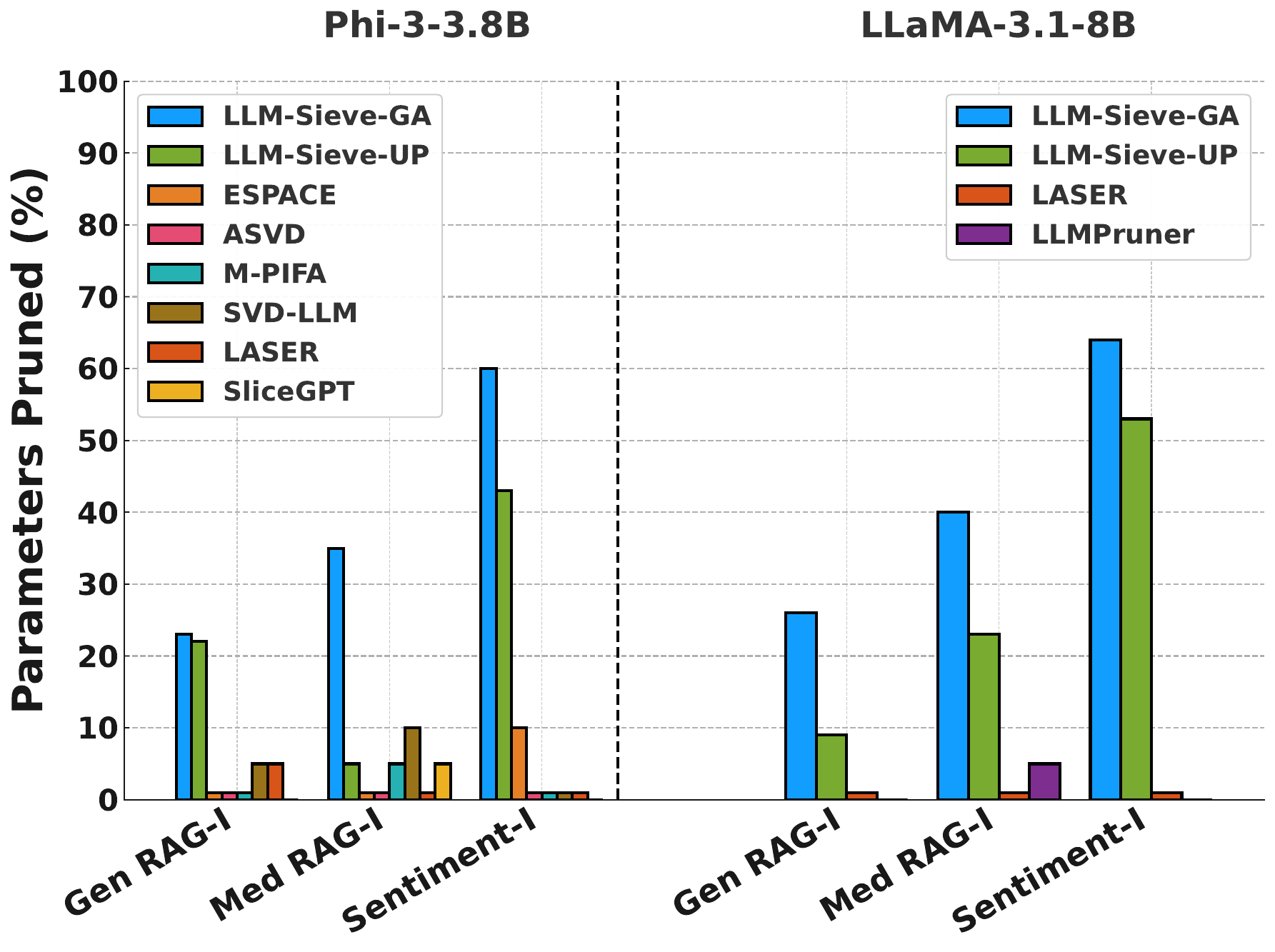}
        \caption{\small{Comparison of LLM-Sieve with State-of-the-art}}
        \label{fig:comparison_results}
    \end{minipage}
    \begin{minipage}[t]{0.64\columnwidth}
    \centering
    \includegraphics[width=\linewidth]{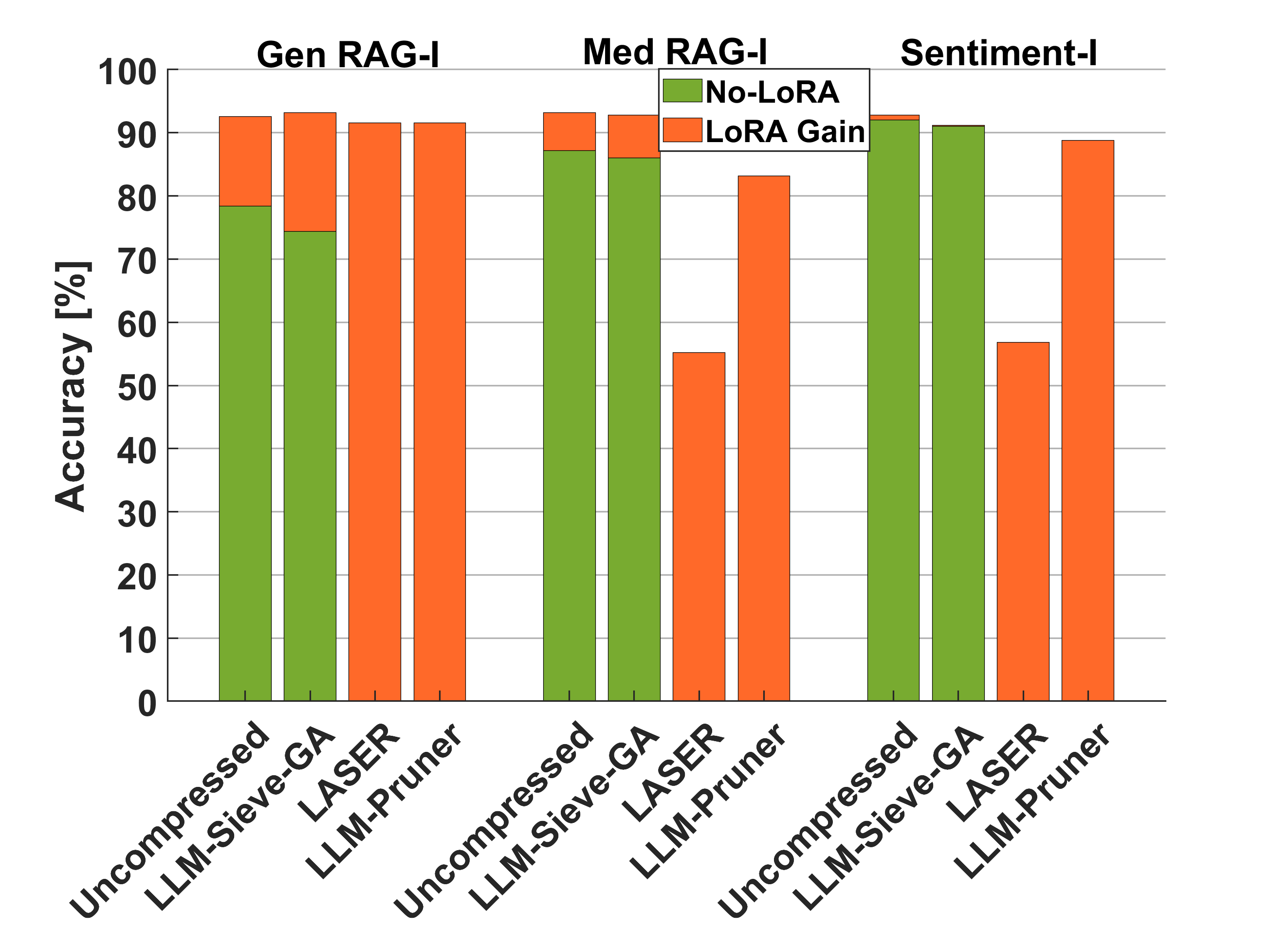}
    \caption{\small Effect of LoRA fine-tuning on LLaMA-3.1-8B.}
    \label{fig:LoRA-SameData}
    \end{minipage}
    \begin{minipage}[t]{0.64\columnwidth}
    \centering
    \includegraphics[width=\linewidth]{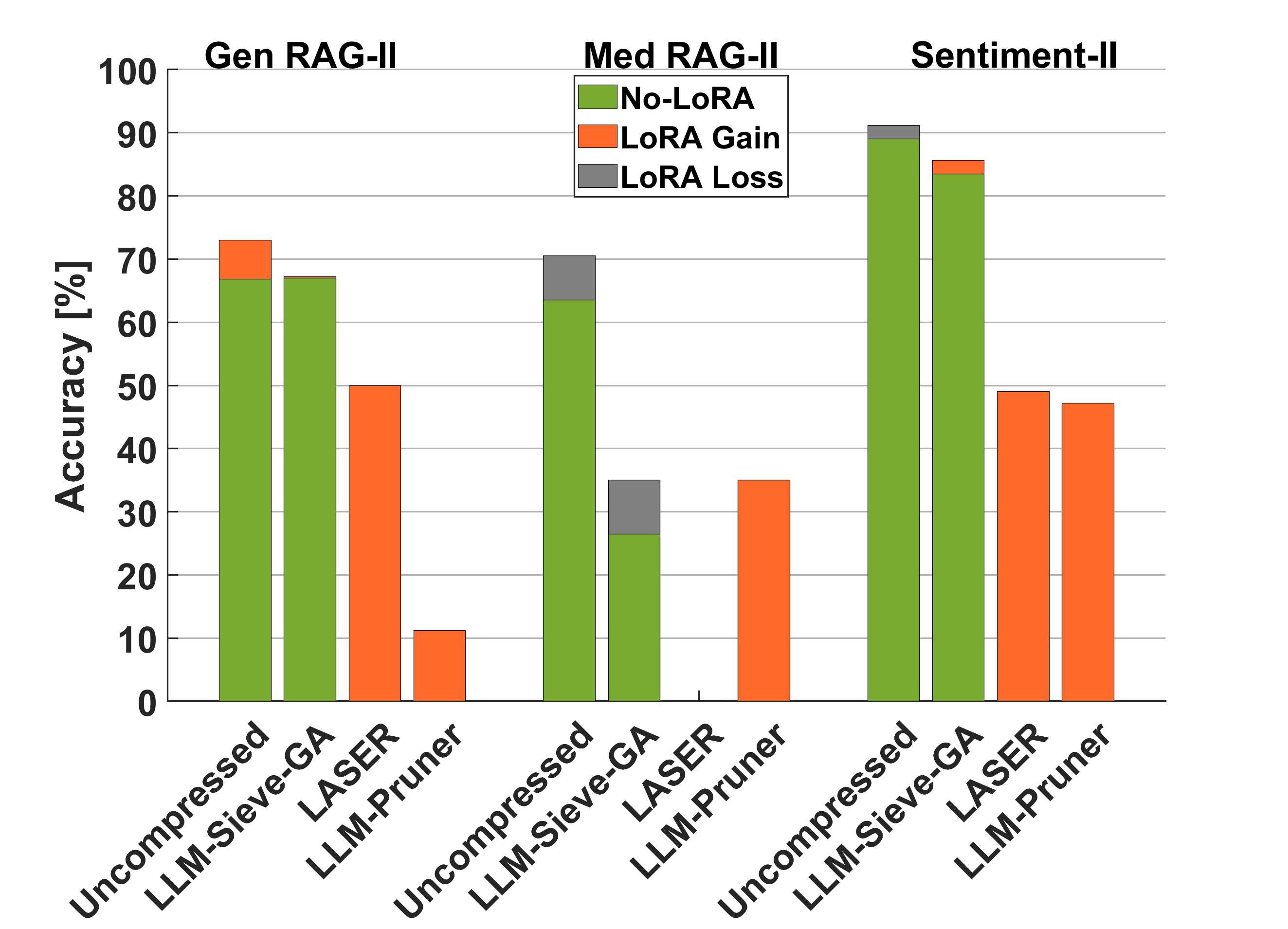}
    \caption{\small Effect of LoRA fine-tuning on unseen datasets for LLaMA-3.1-8B.}
    \label{fig:LoRA-DifferentData}
    \end{minipage}
    \vspace{-0.15in}
\end{figure*}

\begin{table*}[htbp]
\centering
\small
\setlength{\tabcolsep}{25pt}
\renewcommand{\arraystretch}{1.15}
\caption{Accuracy (\%) of different compression methods across tasks on Phi-3-mini.}
\label{tab:comparison_results}
\begin{tabular}{@{}lccc@{}}
\toprule
\textbf{Method (Parameter Pruned \%)} & \textbf{Generic RAG-I} & \textbf{Medical RAG-I} & \textbf{Sentiment-I} \\
\midrule
\textcolor{RowTextGray}{Uncompressed} & \textcolor{RowTextGray}{66} & \textcolor{RowTextGray}{68} & \textcolor{RowTextGray}{93} \\
LLM-Sieve-GA (17 / 35 / 44) & \textbf{64} & \textbf{67} & \textbf{92} \\
ESPACE (10)                 & 20 & 14 & 90 \\
ASVD (10)                   & 18 & 29 & 29 \\
M-PIFA (10)                 & 23 & 41 & 8 \\
SVD-LLM (10)                & 46 & 64 & 3 \\
LASER (10)                  & 30 & 0  & 0 \\
SliceGPT (10)               & 5  & 0  & 0 \\
\bottomrule
\end{tabular}
\end{table*}

\subsection{Comparison to State-of-the-Art Methods}
\label{sec:eval-comparison}
To ensure fairness, we compare LLM-Sieve against state-of-the-art pruning methods under the constraint that accuracy remains within 5\% of the uncompressed model. Since most prior techniques use uniform pruning without a clear way to select pruning levels, we augment all baselines with binary search to identify the highest compression achievable at \(\epsilon \!\le\! 5\%\). Experiments are conducted on Phi-3-3.8B and LLaMA-3.1-8B; among baselines, only LASER and LLM-Pruner provide implementations compatible with LLaMA.  

\noindent
{\bf LLM-Sieve-GA significantly outperforms baselines.}  
As shown in Figure~\ref{fig:comparison_results}, both LLM-Sieve-GA and LLM-Sieve-UP outperform all other techniques: while prior methods prune at most 1–10\% of parameters, LLM-Sieve removes 20–65\% with comparable accuracy.  

\noindent
{\bf Output-aligned non-orthogonal projections outperform SVD/PCA.}  
The fact that LLM-Sieve-UP (without adaptive pruning) still surpasses other methods indicates that output-aligned, non-orthogonal projections preserve task-specific behavior more faithfully than traditional SVD/PCA.  

\noindent
{\bf State-of-the-art fails even at 10\% pruning.}  
Table~\ref{tab:comparison_results} shows that while most baselines collapse at 10\% pruning, LLM-Sieve achieves much higher compression—17\% for Generic RAG-I, 35\% for Medical RAG-I, and 44\% for Sentiment-I—with negligible accuracy loss.

\begin{figure*}[t]
    \centering
    \begin{minipage}[t]{0.64\columnwidth}
        \centering
        \includegraphics[width=\linewidth]{figures/evaluation/CompressionVsAccuracy.png}
        \caption{\small{Accuracy vs. pruning \% with and without 8-bit quantization.}}
        \label{fig:compressionvsaccuracy}
    \end{minipage}
    \hfill
    \begin{minipage}[t]{0.64\columnwidth}
        \centering
        \includegraphics[width=1\textwidth]{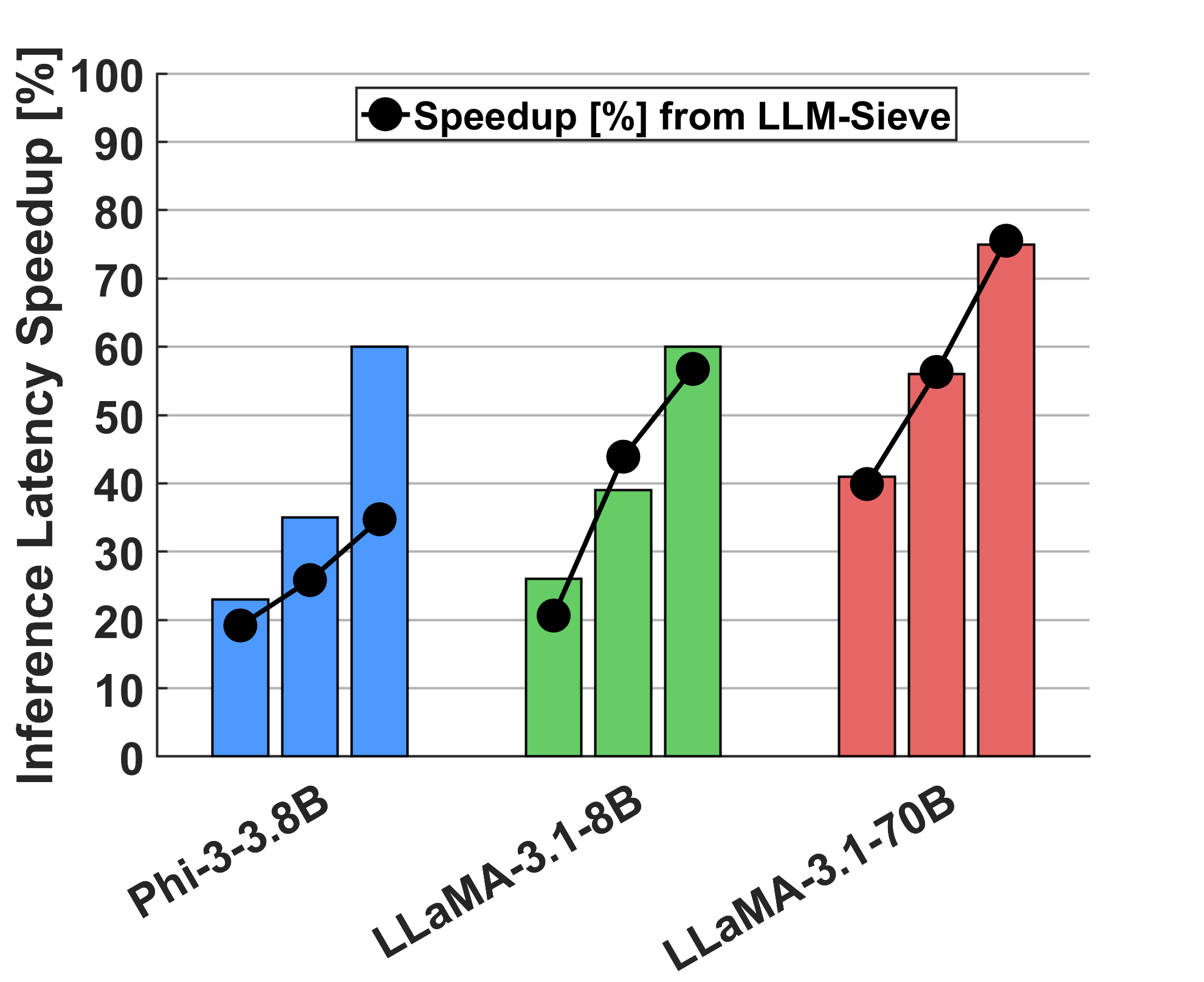}
    \caption{\small {Reduction in parameters vis-a-vis reduction in inference time (speedup for LLM-Sieve).}}
    \label{fig:speedup}
    \end{minipage}
    \hfill
    \begin{minipage}[t]{0.64\columnwidth}
        \centering
        \includegraphics[width=\linewidth]{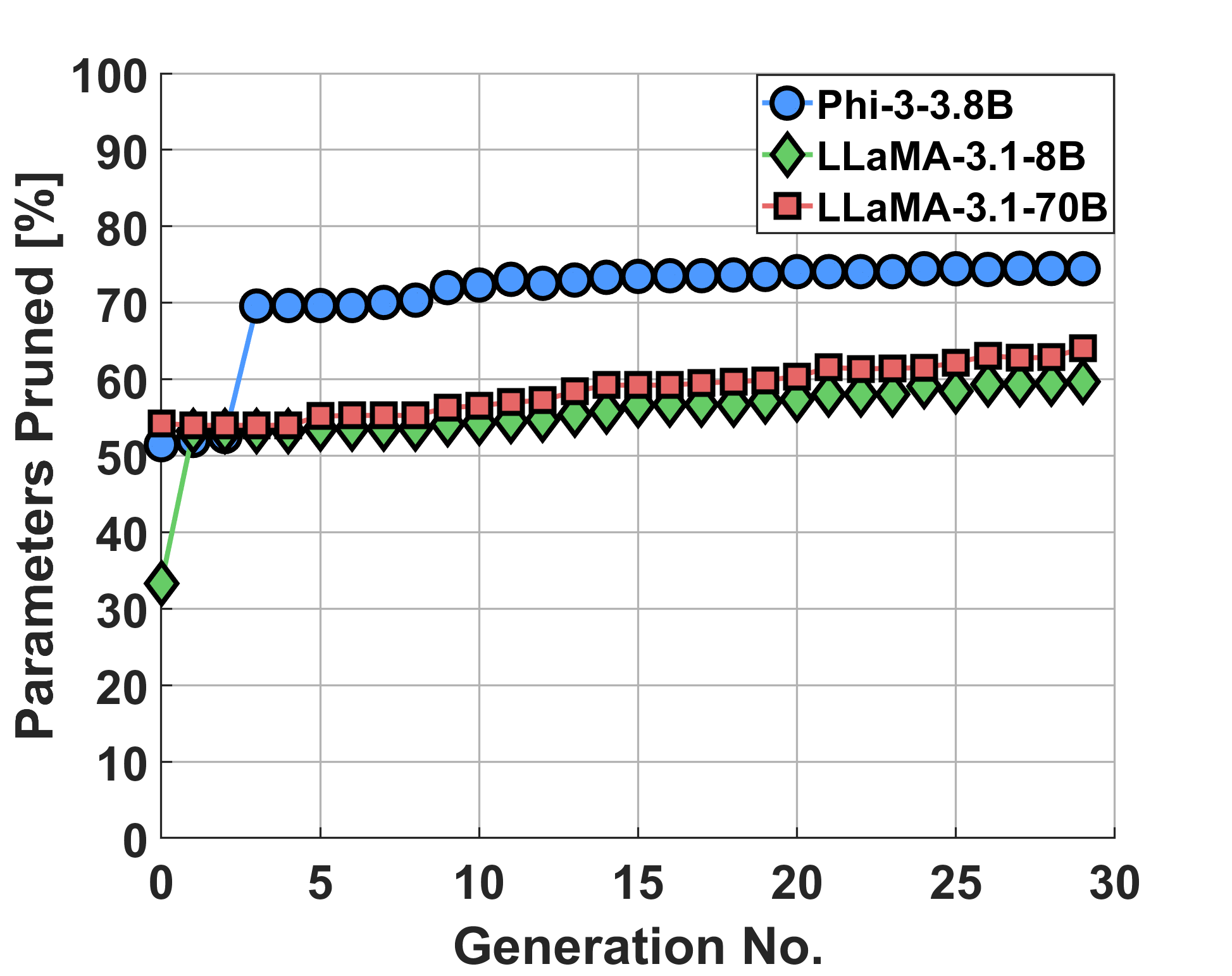}
    \caption{\small{How genetic algorithm parameter reduction evolves across generations.}}
    \label{fig:appendix_ga_convergence}
    \end{minipage}
    \vspace{-0.15in}
\end{figure*}

\subsection{LoRA Fine-Tuning and Cross-Dataset Generalization}
\label{sec:eval-LoRA}
LoRA fine-tuning~\cite{lora} is widely used to specialize LLMs for downstream tasks. While it can improve accuracy, we find that \emph{heavy reliance on LoRA undermines generalization across datasets within the same domain.}  

\textbf{LORA on Same-dataset.} Figure~\ref{fig:LoRA-SameData} shows results when models are fine-tuned on the same dataset used for calibration. For fairness, we prune all methods to the maximum compression level achieved by LLM-Sieve under a 5\% accuracy drop, and restrict comparisons to LASER and LLM-Pruner (the only baselines supporting fine-tuning). LLM-Sieve-GA benefits modestly from LoRA, similar to gains in the uncompressed model across all tasks. By contrast, LASER and LLM-Pruner—near-zero before fine-tuning—jump to 50–90\% accuracy, revealing strong dependence on post-pruning LoRA fine-tuning.  

\textbf{Cross-dataset evaluation.} Figure~\ref{fig:LoRA-DifferentData} and Table~\ref{tab:generalization} evaluate transfer to unseen datasets within each domain. LLM-Sieve maintains high accuracy: no loss on Gen RAG-II, $\sim$7\% loss on Sentiment-II, and only a drop on Med RAG-II. The latter stems from output format mismatch: Med RAG-I expects True/False answers, while Med RAG-II requires multiple-choice responses—yet LoRA-tuned models persisted in generating True/False outputs despite prompts. LASER and LLM-Pruner generalize far worse, reflecting their dependence on LoRA for recovery.

\begin{table}[htbp]
    \centering
    \caption{\small Generalization of LLM-Sieve-GA across datasets for the same task (LLaMA-3.1-8B).}
    \label{tab:generalization}
    \resizebox{\columnwidth}{!}{%
        \begin{tabular}{@{}l l c c@{}}
            \toprule
            \small{Task} & \small{Test Dataset} & \small{Uncompressed} & \small{LLM-Sieve-GA} \\
            & & \small{Accuracy [\%]} & \small{Accuracy [\%]} \\
            \midrule
            \small{General RAG}       & \small{Gen RAG-II}     & \small{67} & \small{67} \\
            \small{Medical Q\&A}      & \small{Med RAG-II}     & \small{71} & \small{35} \\
            \small{Sentiment Analysis}& \small{Sentiment-II}   & \small{91} & \small{84} \\
            \bottomrule
        \end{tabular}%
    }
\end{table}

Overall, LoRA can improve pruned models but at the cost of dataset generalization. In contrast, LLM-Sieve achieves strong cross-dataset robustness without relying on LoRA, suggesting pruning itself—not post-hoc fine-tuning—is the key driver of task sufficiency.

\subsection{Quantization With LLM-Sieve}
\label{sec:quatization}
Quantization reduces memory and latency by encoding weights in lower precision (e.g., 8-bit integers). Figure~\ref{fig:compressionvsaccuracy} shows accuracy–compression curves for LLM-Sieve-GA with and without quantization on Phi-3-3.8B, LLaMA-3.1-8B, and LLaMA-3.1-70B. In all cases, accuracy plateaus until a sharp drop, indicating that many parameters are not task-critical. Importantly, the quantized models (marked “-Q”) track the unquantized ones almost exactly, showing that quantization can be applied after pruning to halve memory and reduce latency with minimal additional loss in accuracy.

\subsection{Reduction in Inference Latency}
\label{sec:speedup}
Figure~\ref{fig:speedup} reports wall-clock inference latency on a microbenchmark across three models and pruning levels, using CUTLASS~\cite{cutlass} to accelerate tensor operations (with matrix dimensions constrained to powers of two). Latency speedup scales nearly linearly with the fraction of parameters removed. The only deviation is for Phi-3-3.8B, where smaller matrices limit GPU pipeline saturation, yielding slightly sub-linear gains. Even so, pruning delivers substantial and consistent performance improvements across all models.

\subsection{Running Time and Tradeoffs for LLM-Sieve}
\label{sec:eval-ga}
LLM-Sieve incurs a one-time cost comparable to LoRA fine-tuning. As shown in Table~\ref{tab:gpu_hours}, pruning itself requires 1–36 GPU hours depending on model size. The main expense is searching for pruning factors: uniform pruning with binary search (Sieve-UP) converges in 2–4 steps, while differentiated pruning with a Genetic Algorithm (GA) converges in 10–15 generations, totaling 144–900 GPU hours (Figure~\ref{fig:appendix_ga_convergence}).  

These results highlight a tradeoff. For smaller models (e.g., Phi-3-3.8B, LLaMA-3.1-8B), the modest accuracy gains from adaptive pruning may not justify its higher cost, making uniform pruning a practical compromise. For larger models (e.g., LLaMA-3.1-70B), however, the substantial accuracy improvements make adaptive pruning essential despite its runtime overhead.

\begin{table}[h]
    \centering
    \caption{\small GPU Hours Spent}
    \label{tab:gpu_hours}
    \resizebox{\columnwidth}{!}{%
        \begin{tabular}{@{}lrrr@{}}
            \toprule
            Model & Pruning & LLM-Sieve-UP & LLM-Sieve-GA \\
            \midrule
            Phi-3-3.8B   & 2  & 5  & 144 \\
            LLaMA-3.1-8B & 3  & 8  & 270 \\
            LLaMA-3.1-70B& 35 & 85 & 891 \\
            \bottomrule
        \end{tabular}%
    }
\end{table}

\section{Conclusion, Limitations \& Future Work}
\label{sec:conclusion}

We introduced LLM-Sieve, a framework that prunes LLMs down to the minimal parameter subset needed to preserve task performance.
By combining joint low-rank projections with a Genetic Algorithm for adaptive pruning, LLM-Sieve enables fine-grained compression tailored to task structure. As a result, it significantly outperforms prior state-of-the-art methods across different models and tasks, and remains compatible with LoRA fine-tuning and quantization—supporting a practical pipeline for efficient, task-adapted LLMs.  

A key limitation is that LLM-Sieve retains the full model architecture, including layer count, which caps compression potential. In contrast, distillation can yield much smaller models (e.g., LLaMA-8B from LLaMA-70B) but at significantly higher cost. Currently, pruning is done per-matrix to avoid backpropagation through non-linearities, and the Genetic Algorithm, while effective, may be replaced by faster pruning-factor search methods. Beyond efficiency, pruning also serves as a probe into how knowledge is organized across layers and matrices, suggesting architectural implications—future models may benefit from non-uniform capacity allocation, leading to designs that are both leaner and more transparent in function.

\newpage

\bibliography{references}

\begin{thebibliography}{41}
\providecommand{\natexlab}[1]{#1}
\providecommand{\url}[1]{\texttt{#1}}
\expandafter\ifx\csname urlstyle\endcsname\relax
  \providecommand{\doi}[1]{doi: #1}\else
  \providecommand{\doi}{doi: \begingroup \urlstyle{rm}\Url}\fi

\bibitem[Zafrir et~al.(2021)Zafrir, Larey, Boudoukh, Shen, and Wasserblat]{zafrir2021prune}
Ofir Zafrir, Ariel Larey, Guy Boudoukh, Haihao Shen, and Moshe Wasserblat.
\newblock Prune once for all: Sparse pre-trained language models.
\newblock \emph{arXiv preprint arXiv:2111.05754}, 2021.

\bibitem[Frankle and Carbin(2018)]{frankle2018lottery}
Jonathan Frankle and Michael Carbin.
\newblock The lottery ticket hypothesis: Finding sparse, trainable neural networks.
\newblock \emph{arXiv preprint arXiv:1803.03635}, 2018.

\bibitem[Frantar and Alistarh(2023)]{sparsegpt}
Elias Frantar and Dan Alistarh.
\newblock Sparsegpt: Massive language models can be accurately pruned in one-shot.
\newblock In \emph{International Conference on Machine Learning}, pages 10323--10337. PMLR, 2023.

\bibitem[Sun et~al.(2023)Sun, Liu, Bair, and Kolter]{wanda}
Mingjie Sun, Zhuang Liu, Anna Bair, and J~Zico Kolter.
\newblock A simple and effective pruning approach for large language models.
\newblock \emph{arXiv preprint arXiv:2306.11695}, 2023.

\bibitem[Ma et~al.(2023)Ma, Fang, and Wang]{llmpruner}
Xinyin Ma, Gongfan Fang, and Xinchao Wang.
\newblock Llm-pruner: On the structural pruning of large language models.
\newblock \emph{Advances in neural information processing systems}, 36:\penalty0 21702--21720, 2023.

\bibitem[van~der Ouderaa et~al.(2023)van~der Ouderaa, Nagel, Van~Baalen, Asano, and Blankevoort]{llmsurgeon}
Tycho~FA van~der Ouderaa, Markus Nagel, Mart Van~Baalen, Yuki~M Asano, and Tijmen Blankevoort.
\newblock The llm surgeon.
\newblock \emph{arXiv preprint arXiv:2312.17244}, 2023.

\bibitem[Wang et~al.(2019)Wang, Grosse, Fidler, and Zhang]{eigendamage}
Chaoqi Wang, Roger Grosse, Sanja Fidler, and Guodong Zhang.
\newblock Eigendamage: Structured pruning in the kronecker-factored eigenbasis.
\newblock In \emph{International conference on machine learning}, pages 6566--6575. PMLR, 2019.

\bibitem[Noach and Goldberg(2020)]{noach2020compressing}
Matan~Ben Noach and Yoav Goldberg.
\newblock Compressing pre-trained language models by matrix decomposition.
\newblock In \emph{Proceedings of the 1st Conference of the Asia-Pacific Chapter of the Association for Computational Linguistics and the 10th International Joint Conference on Natural Language Processing}, pages 884--889, 2020.

\bibitem[Ashkboos et~al.(2024{\natexlab{a}})Ashkboos, Croci, Nascimento, Hoefler, and Hensman]{slicegpt}
Saleh Ashkboos, Maximilian~L Croci, Marcelo Gennari~do Nascimento, Torsten Hoefler, and James Hensman.
\newblock Slicegpt: Compress large language models by deleting rows and columns.
\newblock \emph{arXiv preprint arXiv:2401.15024}, 2024{\natexlab{a}}.

\bibitem[Sakr and Khailany(2024)]{espace}
Charbel Sakr and Brucek Khailany.
\newblock Espace: Dimensionality reduction of activations for model compression.
\newblock \emph{Advances in Neural Information Processing Systems}, 37:\penalty0 17489--17517, 2024.

\bibitem[Sharma et~al.(2023)Sharma, Ash, and Misra]{laser}
Pratyusha Sharma, Jordan~T Ash, and Dipendra Misra.
\newblock The truth is in there: Improving reasoning in language models with layer-selective rank reduction.
\newblock \emph{arXiv preprint arXiv:2312.13558}, 2023.

\bibitem[Wang et~al.(2024)Wang, Zheng, Wan, and Zhang]{svdllm}
Xin Wang, Yu~Zheng, Zhongwei Wan, and Mi~Zhang.
\newblock Svd-llm: Truncation-aware singular value decomposition for large language model compression.
\newblock \emph{arXiv preprint arXiv:2403.07378}, 2024.

\bibitem[Yuan et~al.(2023)Yuan, Shang, Song, Wu, Yan, and Sun]{asvd}
Zhihang Yuan, Yuzhang Shang, Yue Song, Qiang Wu, Yan Yan, and Guangyu Sun.
\newblock Asvd: Activation-aware singular value decomposition for compressing large language models.
\newblock \emph{arXiv preprint arXiv:2312.05821}, 2023.

\bibitem[Zhao et~al.(2025)Zhao, Zhang, and Cannistraci]{mpifa}
Jialin Zhao, Yingtao Zhang, and Carlo~Vittorio Cannistraci.
\newblock Pivoting factorization: A compact meta low-rank representation of sparsity for efficient inference in large language models.
\newblock \emph{arXiv preprint arXiv:2501.19090}, 2025.

\bibitem[Frantar et~al.(2022)Frantar, Ashkboos, Hoefler, and Alistarh]{gptq}
Elias Frantar, Saleh Ashkboos, Torsten Hoefler, and Dan Alistarh.
\newblock Gptq: Accurate post-training quantization for generative pre-trained transformers.
\newblock \emph{arXiv preprint arXiv:2210.17323}, 2022.

\bibitem[Xiao et~al.(2023)Xiao, Lin, Seznec, Wu, Demouth, and Han]{smoothquant}
Guangxuan Xiao, Ji~Lin, Mickael Seznec, Hao Wu, Julien Demouth, and Song Han.
\newblock Smoothquant: Accurate and efficient post-training quantization for large language models.
\newblock In \emph{International Conference on Machine Learning}, pages 38087--38099. PMLR, 2023.

\bibitem[Ashkboos et~al.(2024{\natexlab{b}})Ashkboos, Mohtashami, Croci, Li, Cameron, Jaggi, Alistarh, Hoefler, and Hensman]{quarot}
Saleh Ashkboos, Amirkeivan Mohtashami, Maximilian Croci, Bo~Li, Pashmina Cameron, Martin Jaggi, Dan Alistarh, Torsten Hoefler, and James Hensman.
\newblock Quarot: Outlier-free 4-bit inference in rotated llms.
\newblock \emph{Advances in Neural Information Processing Systems}, 37:\penalty0 100213--100240, 2024{\natexlab{b}}.

\bibitem[Liu et~al.(2024)Liu, Zhao, Fedorov, Soran, Choudhary, Krishnamoorthi, Chandra, Tian, and Blankevoort]{spinquant}
Zechun Liu, Changsheng Zhao, Igor Fedorov, Bilge Soran, Dhruv Choudhary, Raghuraman Krishnamoorthi, Vikas Chandra, Yuandong Tian, and Tijmen Blankevoort.
\newblock Spinquant: Llm quantization with learned rotations.
\newblock \emph{arXiv preprint arXiv:2405.16406}, 2024.

\bibitem[Gholami et~al.(2022)Gholami, Kim, Dong, Yao, Mahoney, and Keutzer]{gholami2022survey}
Amir Gholami, Sehoon Kim, Zhen Dong, Zhewei Yao, Michael~W Mahoney, and Kurt Keutzer.
\newblock A survey of quantization methods for efficient neural network inference.
\newblock In \emph{Low-Power Computer Vision}, pages 291--326. Chapman and Hall/CRC, 2022.

\bibitem[Hoefler et~al.(2021)Hoefler, Alistarh, Ben-Nun, Dryden, and Peste]{hoefler2021sparsity}
Torsten Hoefler, Dan Alistarh, Tal Ben-Nun, Nikoli Dryden, and Alexandra Peste.
\newblock Sparsity in deep learning: Pruning and growth for efficient inference and training in neural networks.
\newblock \emph{Journal of Machine Learning Research}, 22\penalty0 (241):\penalty0 1--124, 2021.

\bibitem[Hinton(2015)]{hinton2015distilling}
Geoffrey Hinton.
\newblock Distilling the knowledge in a neural network.
\newblock \emph{arXiv preprint arXiv:1503.02531}, 2015.

\bibitem[Vaswani(2017)]{attention}
A~Vaswani.
\newblock Attention is all you need.
\newblock \emph{Advances in Neural Information Processing Systems}, 2017.

\bibitem[Ba(2016)]{layernorm}
Jimmy~Lei Ba.
\newblock Layer normalization.
\newblock \emph{arXiv preprint arXiv:1607.06450}, 2016.

\bibitem[Zhang and Sennrich(2019)]{rmsnorm}
Biao Zhang and Rico Sennrich.
\newblock Root mean square layer normalization.
\newblock \emph{Advances in Neural Information Processing Systems}, 32, 2019.

\bibitem[Wang et~al.(2018)Wang, Singh, Michael, Hill, Levy, and Bowman]{glue}
Alex Wang, Amanpreet Singh, Julian Michael, Felix Hill, Omer Levy, and Samuel Bowman.
\newblock {GLUE}: A multi-task benchmark and analysis platform for natural language understanding.
\newblock In \emph{Proceedings of the 2018 EMNLP Workshop BlackboxNLP}, pages 353--355, Brussels, Belgium, November 2018. Association for Computational Linguistics.
\newblock \doi{10.18653/v1/W18-5446}.
\newblock URL \url{https://aclanthology.org/W18-5446/}.

\bibitem[Bommasani et~al.(2023)Bommasani, Hudson, et~al.]{helm}
Rishi Bommasani, Drew~A. Hudson, et~al.
\newblock Holistic evaluation of language models (helm).
\newblock \emph{New York Academy of Sciences}, 2023.
\newblock URL \url{https://crfm.stanford.edu/helm/}.
\newblock Living benchmark for LM evaluation.

\bibitem[Bai et~al.(2024)Bai, Lv, Zhang, Lyu, Tang, Huang, Du, Liu, Zeng, Hou, Dong, Tang, and Li]{longbench}
Yushi Bai, Xin Lv, Jiajie Zhang, Hongchang Lyu, Jiankai Tang, Zhidian Huang, Zhengxiao Du, Xiao Liu, Aohan Zeng, Lei Hou, Yuxiao Dong, Jie Tang, and Juanzi Li.
\newblock Longbench: A bilingual, multi-task benchmark for long context understanding.
\newblock In \emph{Proceedings of ACL (Long/Multilingual Track)}, 2024.
\newblock URL \url{https://aclanthology.org/2024.acl-long.172/}.
\newblock Also arXiv 2308.14508.

\bibitem[authors)(2020)]{blurb}
TBD (Microsoft /~BLURB authors).
\newblock Blurb: Biomedical language understanding \& reasoning benchmark.
\newblock \url{https://microsoft.github.io/BLURB/}, 2020.
\newblock Biomedical NLP benchmark, 13 datasets over 6 tasks.

\bibitem[Singhal et~al.(2023)Singhal, Azizi, Tu, et~al.]{multimedqa}
Karan Singhal, Shekoofeh Azizi, Tao Tu, et~al.
\newblock Large language models encode clinical knowledge: the multimedqa benchmark.
\newblock \emph{Nature / preprint}, 2023.
\newblock URL \url{https://pubmed.ncbi.nlm.nih.gov/37438534/}.

\bibitem[Petroni et~al.(2021)Petroni, Piktus, Fan, Lewis, Yazdani, De~Cao, Thorne, Jernite, Karpukhin, Maillard, Plachouras, and Rocktäschel]{kilt}
Fabio Petroni, Aleksandra Piktus, Angela Fan, Patrick Lewis, Majid Yazdani, Nicola De~Cao, James Thorne, Yacine Jernite, Vladimir Karpukhin, Jean Maillard, Vassilis Plachouras, and Tim Rocktäschel.
\newblock Kilt: A benchmark for knowledge intensive language tasks.
\newblock In \emph{Proceedings of the 2021 NAACL: Human Language Technologies}, 2021.
\newblock URL \url{https://aclanthology.org/2021.naacl-main.235/}.
\newblock Also known as KILT: knowledge-intensive language tasks.

\bibitem[Yang et~al.(2018)Yang, Qi, Zhang, Bengio, Cohen, Salakhutdinov, and Manning]{hotpotqa}
Zhilin Yang, Peng Qi, Saizheng Zhang, Yoshua Bengio, William~W Cohen, Ruslan Salakhutdinov, and Christopher~D Manning.
\newblock Hotpotqa: A dataset for diverse, explainable multi-hop question answering.
\newblock \emph{arXiv preprint arXiv:1809.09600}, 2018.

\bibitem[nat(2025)]{naturalqa}
Natural quesetions - short.
\newblock \url{https://huggingface.co/datasets/cjlovering/natural-questions-short}, 2025.
\newblock [Accessed 16-05-2025].

\bibitem[Jin et~al.(2019)Jin, Dhingra, Liu, Cohen, and Lu]{pubmedqa}
Qiao Jin, Bhuwan Dhingra, Zhengping Liu, William~W Cohen, and Xinghua Lu.
\newblock Pubmedqa: A dataset for biomedical research question answering.
\newblock \emph{arXiv preprint arXiv:1909.06146}, 2019.

\bibitem[med(2022)]{medmcqa}
Medmcqa.
\newblock \url{https://huggingface.co/datasets/openlifescienceai/medmcqa}, 2022.
\newblock [Accessed 16-05-2025].

\bibitem[Maas et~al.(2011)Maas, Daly, Pham, Huang, Ng, and Potts]{imdb}
Andrew Maas, Raymond~E Daly, Peter~T Pham, Dan Huang, Andrew~Y Ng, and Christopher Potts.
\newblock Learning word vectors for sentiment analysis.
\newblock In \emph{Proceedings of the 49th annual meeting of the association for computational linguistics: Human language technologies}, pages 142--150, 2011.

\bibitem[sst(2013)]{sst2}
Stanford sentiment treebank.
\newblock \url{https://huggingface.co/datasets/stanfordnlp/sst}, 2013.
\newblock [Accessed 16-05-2025].

\bibitem[Zheng et~al.(2023)Zheng, Chiang, Sheng, Zhuang, Wu, Zhuang, Lin, Li, Li, Xing, et~al.]{judge}
Lianmin Zheng, Wei-Lin Chiang, Ying Sheng, Siyuan Zhuang, Zhanghao Wu, Yonghao Zhuang, Zi~Lin, Zhuohan Li, Dacheng Li, Eric Xing, et~al.
\newblock Judging llm-as-a-judge with mt-bench and chatbot arena.
\newblock \emph{Advances in Neural Information Processing Systems}, 36:\penalty0 46595--46623, 2023.

\bibitem[Rijsbergen(1979)]{f1score}
C.~J.~Van Rijsbergen.
\newblock \emph{Information Retrieval}.
\newblock Butterworth-Heinemann, 2nd edition, 1979.

\bibitem[Kwon et~al.(2023)Kwon, Li, Zhuang, Sheng, Zheng, Yu, Gonzalez, Zhang, and Stoica]{vllm}
Woosuk Kwon, Zhuohan Li, Siyuan Zhuang, Ying Sheng, Lianmin Zheng, Cody~Hao Yu, Joseph Gonzalez, Hao Zhang, and Ion Stoica.
\newblock Efficient memory management for large language model serving with pagedattention.
\newblock In \emph{Proceedings of the 29th Symposium on Operating Systems Principles}, pages 611--626, 2023.

\bibitem[Hu et~al.(2021)Hu, Shen, Wallis, Allen-Zhu, Li, Wang, Wang, and Chen]{lora}
Edward~J Hu, Yelong Shen, Phillip Wallis, Zeyuan Allen-Zhu, Yuanzhi Li, Shean Wang, Lu~Wang, and Weizhu Chen.
\newblock Lora: Low-rank adaptation of large language models.
\newblock \emph{arXiv preprint arXiv:2106.09685}, 2021.

\bibitem[cut(2025)]{cutlass}
{CUTLASS}.
\newblock \url{https://docs.nvidia.com/cutlass/}, 2025.
\newblock [Accessed 16-05-2025].

\end{thebibliography}
\bibliographystyle{unsrtnat}

\clearpage

\appendix
\thispagestyle{empty}

\onecolumn
\section{Appendix}

\subsection{System Prompts for Tasks \& Evaluation.}
\label{appendix:prompts}
Table \ref{tab:system_prompts} summarizes the system prompts we use both during task-specific evaluations and when employing GPT-4o-as-a-judge. For the tasks, the prompts vary in specificity: Generic RAG has no system prompt, while Medical RAG and Sentiment Analysis are guided by task-specific instructions. In contrast, GPT-4o-as-a-judge uses a carefully designed prompt that enforces strict binary evaluation of model answers against the ground truth, ensuring consistency and reproducibility in automatic scoring.

\begin{table}[h]
\centering
\caption{System prompts used for task-specific evaluations and for GPT-4o-as-a-judge scoring.}
\label{tab:system_prompts}
\renewcommand{\arraystretch}{1.2}
\begin{tabularx}{\columnwidth}{@{}lX@{}}
\toprule
\multicolumn{2}{@{}l}{\textbf{Task Prompts}} \\
\midrule
Generic RAG & None \\
Medical RAG & ``Answer the following question with `yes'', `no'. Provide a reasoning for your answer.'' \\
Sentiment Analysis & ``You are an expert at sentiment analysis. Provide a classification for this text, 0 if it is negative or 1 if positive. Do not write anything else.'' \\
\toprule
\multicolumn{2}{@{}l}{\textbf{Evaluation Prompt}} \\
\midrule
GPT-4o-as-a-judge & ``Check whether the model answer for this prompt is correct compared to the ground truth. If you can't find an answer, and only see the supporting facts and question, then consider it incorrect. The prompt is located between the \texttt{<prompt>} and \texttt{</prompt>} tags. The ground truth is located between the \texttt{<groundtruth>} and \texttt{</groundtruth>} tags. The model answer is located between the \texttt{<answer>} and \texttt{</answer>} tags. Do not include anything else in your response. Only the word `correct' or `incorrect'.'' \\
\bottomrule
\end{tabularx}
\end{table}

\subsection{Calibration Dataset Sensitivity.}
\label{appendix:sensitivity}
\begin{figure}[h!]
    \centering
    \includegraphics[width=1\textwidth]{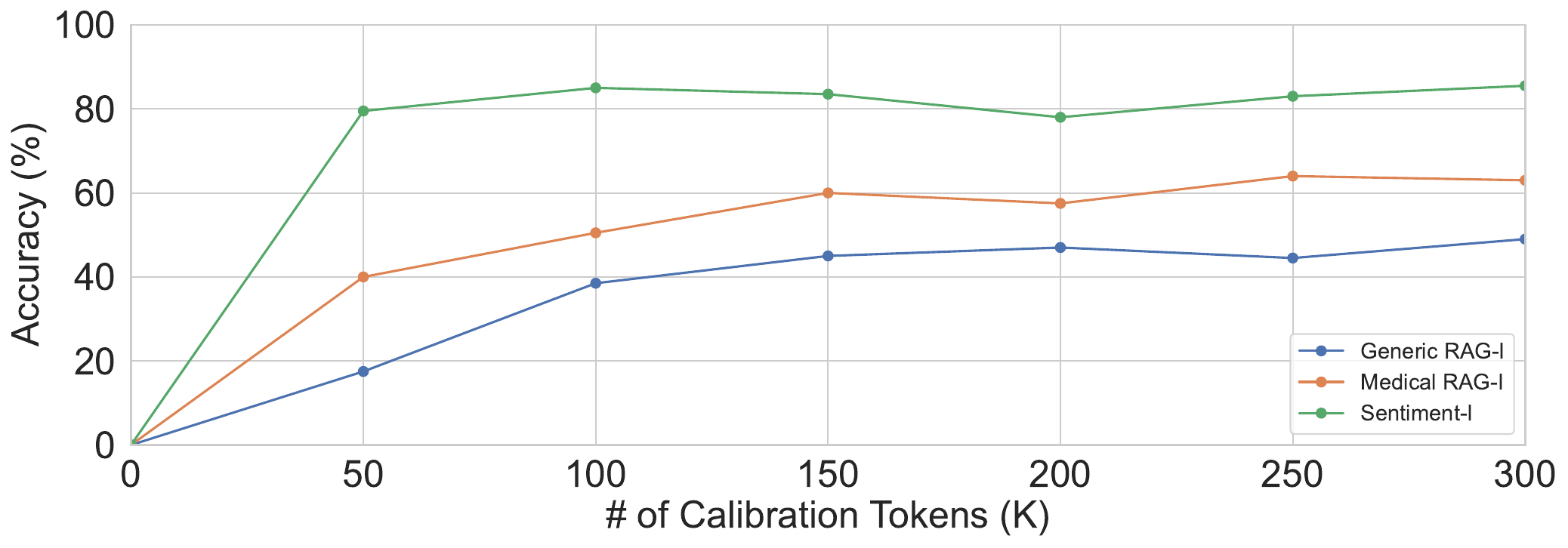}
    \caption{Sensitivity to different calibration dataset sizes for Phi-3-mini. The accuracy benefits start to plateau after 150K tokens.}
    \label{fig:appendix_dataset_sensitivity}
\end{figure}

To assess the impact of calibration dataset size, we ran a sensitivity analysis on Phi-3-mini for all our tasks using LLM-Sieve configured with uniform pruning. As seen in Fig. \ref{fig:appendix_dataset_sensitivity}, we observe that accuracy plateaus after approximately 150K tokens. Based on this, we use a fixed calibration set of 200K tokens for all our pruning experiments.

\subsection{Compressibility of different matrix types.}
\label{appendix:pattern_types}
\begin{figure}[h!]
    \centering
    \begin{subfigure}[b]{0.3\textwidth}
        \centering
        \includegraphics[width=\textwidth]{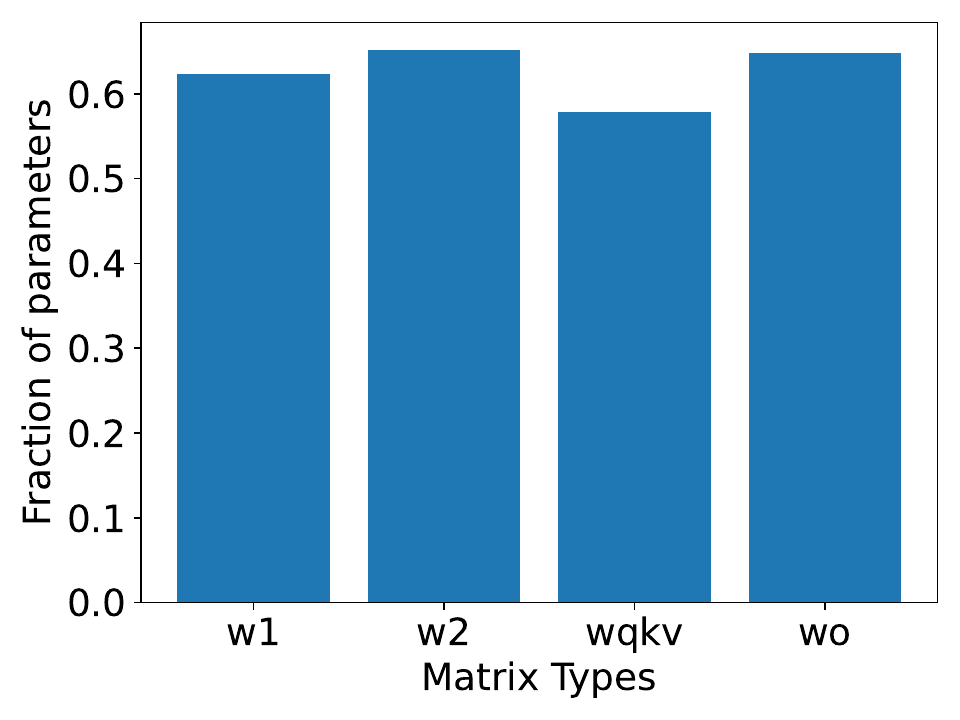}
        \caption{Phi-3-mini+GenRAG-I.}
        \label{fig:phi-mini-averages_by_type-hotpot}
    \end{subfigure}
    \hfill
    \begin{subfigure}[b]{0.3\textwidth}
        \centering
        \includegraphics[width=\textwidth]{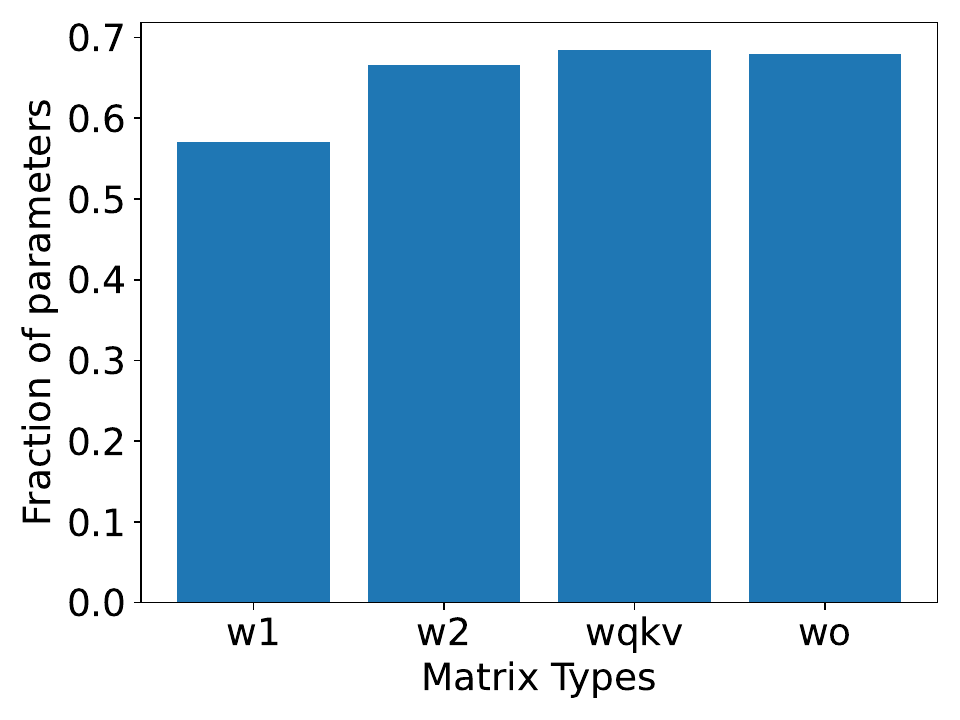}
        \caption{LLaMA-3.1-8B+GenRAG-I.}
        \label{fig:llama-8b-averages_by_type-hotpot}
    \end{subfigure}
    \hfill
    \begin{subfigure}[b]{0.3\textwidth}
        \centering
        \includegraphics[width=\textwidth]{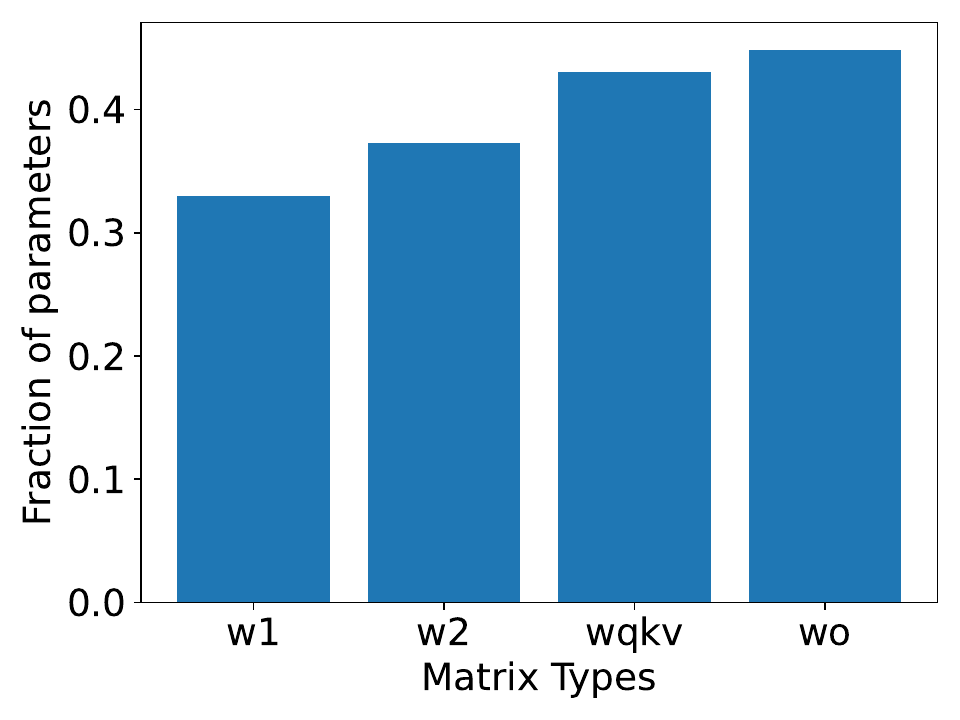}
        \caption{LLaMA-3.1-70B+GenRAG-I.}
        \label{fig:llama-70b-averages_by_type-hotpot}
    \end{subfigure}
    \vskip\baselineskip
      \begin{subfigure}[b]{0.3\textwidth}
        \centering
        \includegraphics[width=\textwidth]{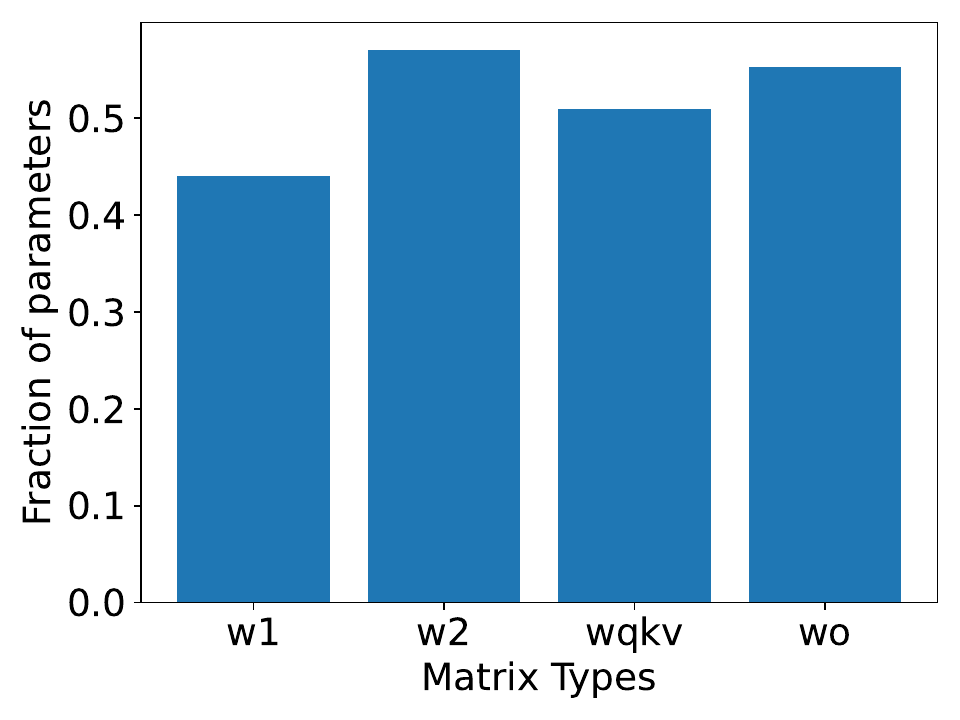}
        \caption{Phi-3-mini+MedRAG-I.}
        \label{fig:phi-mini-averages_by_type-pubmed}
    \end{subfigure}
    \hfill
    \begin{subfigure}[b]{0.3\textwidth}
        \centering
        \includegraphics[width=\textwidth]{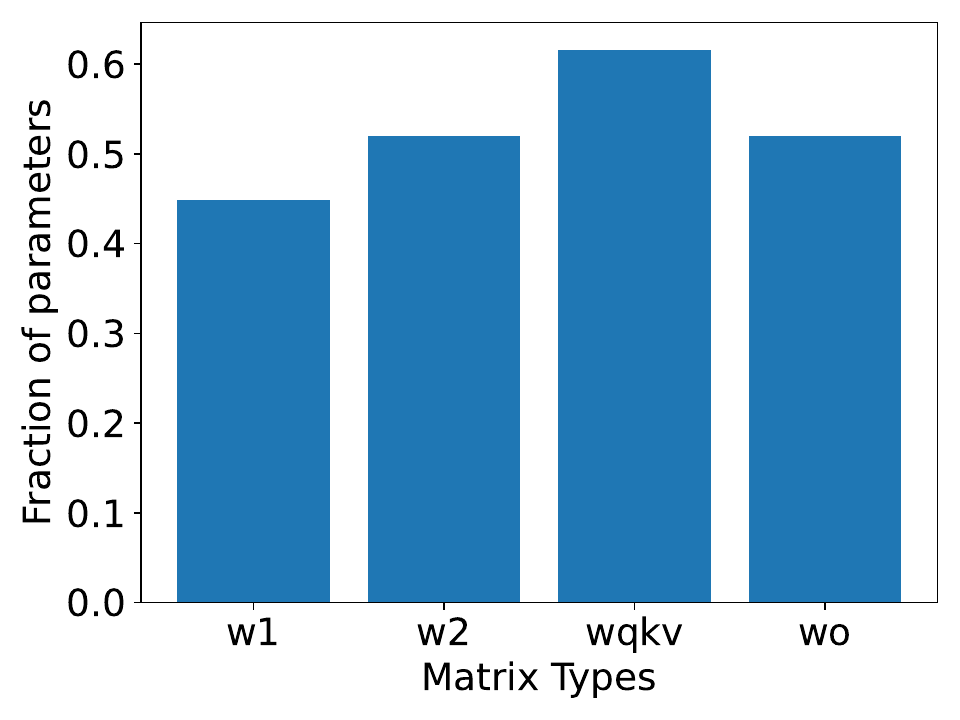}
        \caption{LLaMA-3.1-8B+MedRAG-I.}
        \label{fig:llama-8b-averages_by_type-pubmed}
    \end{subfigure}
    \hfill
    \begin{subfigure}[b]{0.3\textwidth}
        \centering
        \includegraphics[width=\textwidth]{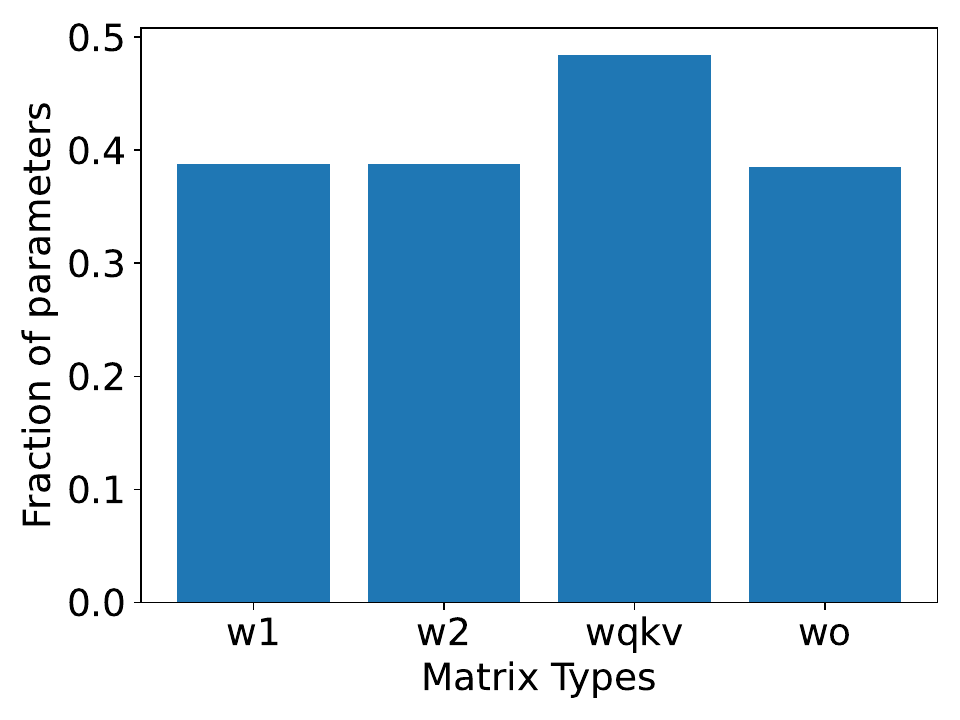}
        \caption{LLaMA-3.1-70B+MedRAG-I.}
        \label{fig:llama-70b-averages_by_type-pubmed}
    \end{subfigure}
    \vskip\baselineskip
    \begin{subfigure}[b]{0.3\textwidth}
        \centering
        \includegraphics[width=\textwidth]{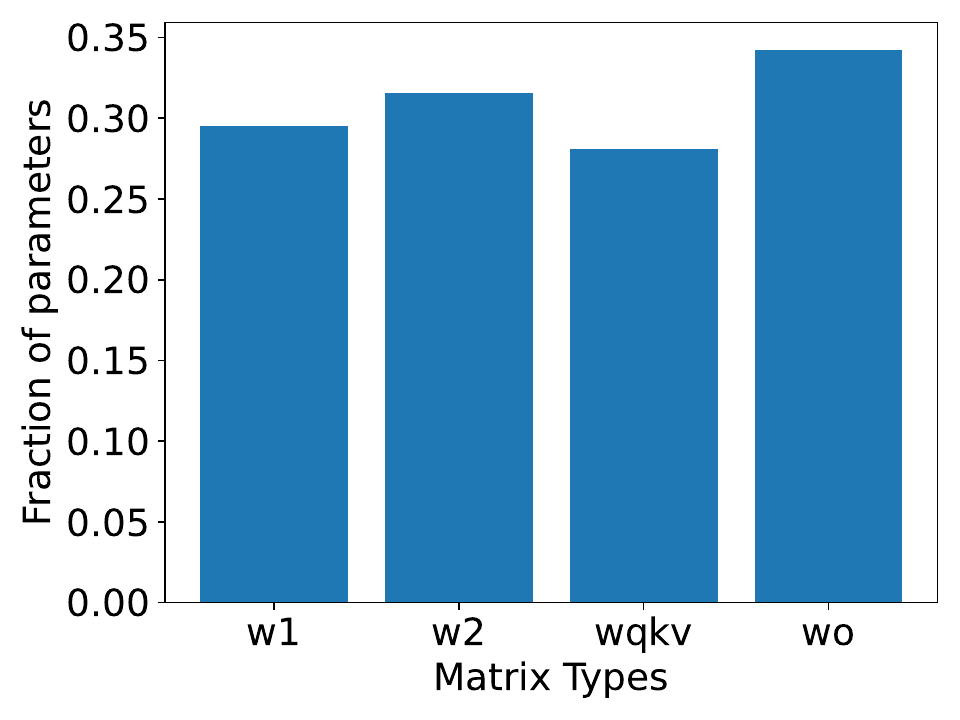}
        \caption{Phi-3-mini+Sentiment-I.}
        \label{fig:phi-mini-averages_by_type-imdb}
    \end{subfigure}
    \hfill
    \begin{subfigure}[b]{0.3\textwidth}
        \centering
        \includegraphics[width=\textwidth]{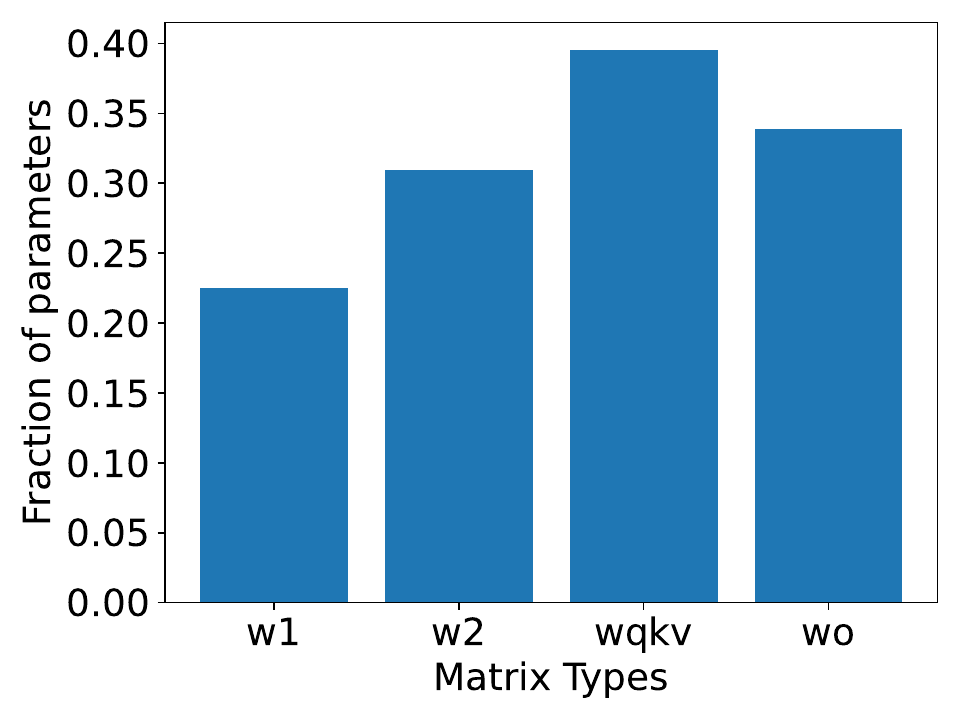}
        \caption{LLaMA-3.1-8B+Sentiment-I.}
        \label{fig:llama-8b-averages_by_type-imdb}
    \end{subfigure}
    \hfill
    \begin{subfigure}[b]{0.3\textwidth}
        \centering
        \includegraphics[width=\textwidth]{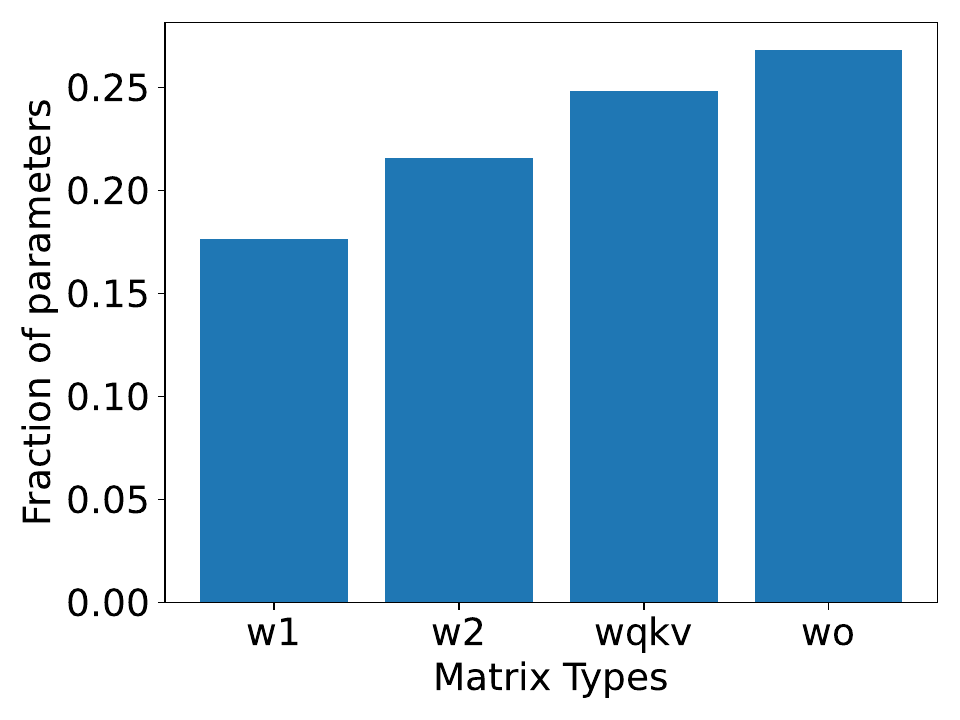}
        \caption{LLaMA-3.1-70B+Sentiment-I.}
        \label{fig:llama-70b-averages_by_type-imdb}
    \end{subfigure}
    \caption{Fraction of remaining parameters for each matrix type found using our Genetic Algorithm search (lower values = higher compression).}
    \label{fig:appendix_genetic_algorithm_by_type}
\end{figure}

In our genetic algorithm (GA) hyperparameter search, we observed that different matrix types within the feedforward and attention components exhibit varying degrees of compressibility. Figure~\ref{fig:appendix_genetic_algorithm_by_type} presents the average fraction of parameters retained per matrix type, averaged across all layers, for the best-performing GA configuration. These results cover all model and dataset combinations. We find that for LLaMA-3.1 8B and 70B, feedforward matrices such as $W_1$ and $W_2$ are generally more compressible than their attention counterparts. This pattern is not observed in Phi-3-mini.

\clearpage

\subsection{Compressibility of different Layers.}
\label{appendix:pattern_layers}
\begin{figure}[h!]
    \begin{subfigure}[b]{0.3\textwidth}
     \centering
    \includegraphics[width=1\textwidth]{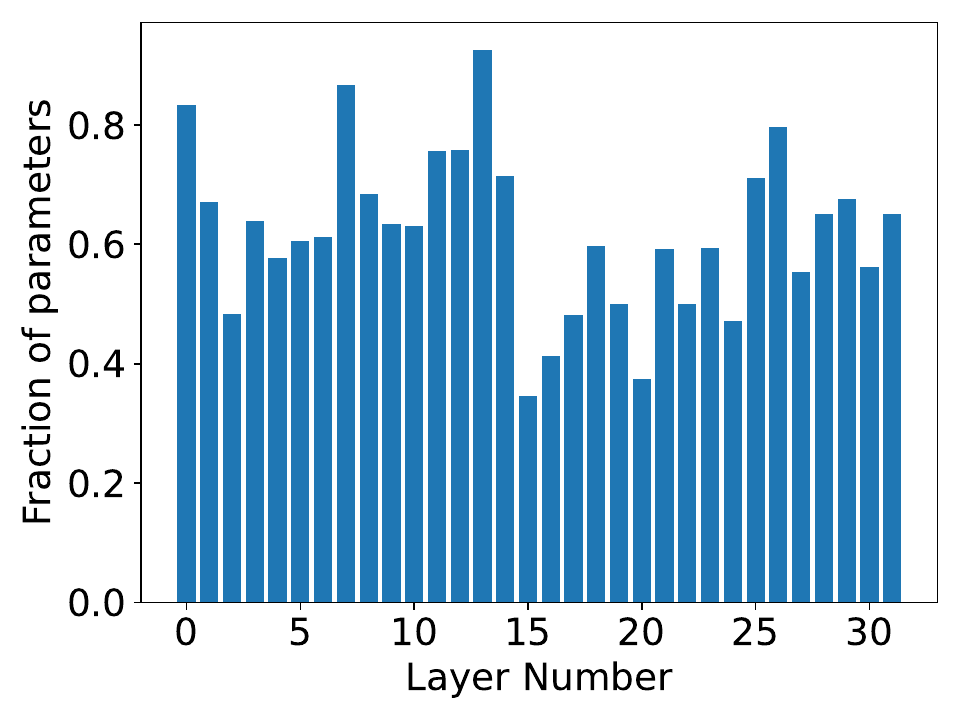}
       \caption{Phi-3-mini+GenRAG-I.}
        \label{fig:phi-mini-averages_by_layer-hotpot}
    \end{subfigure}
    \hfill
    \begin{subfigure}[b]{0.3\textwidth}
     \centering
    \includegraphics[width=1\textwidth]{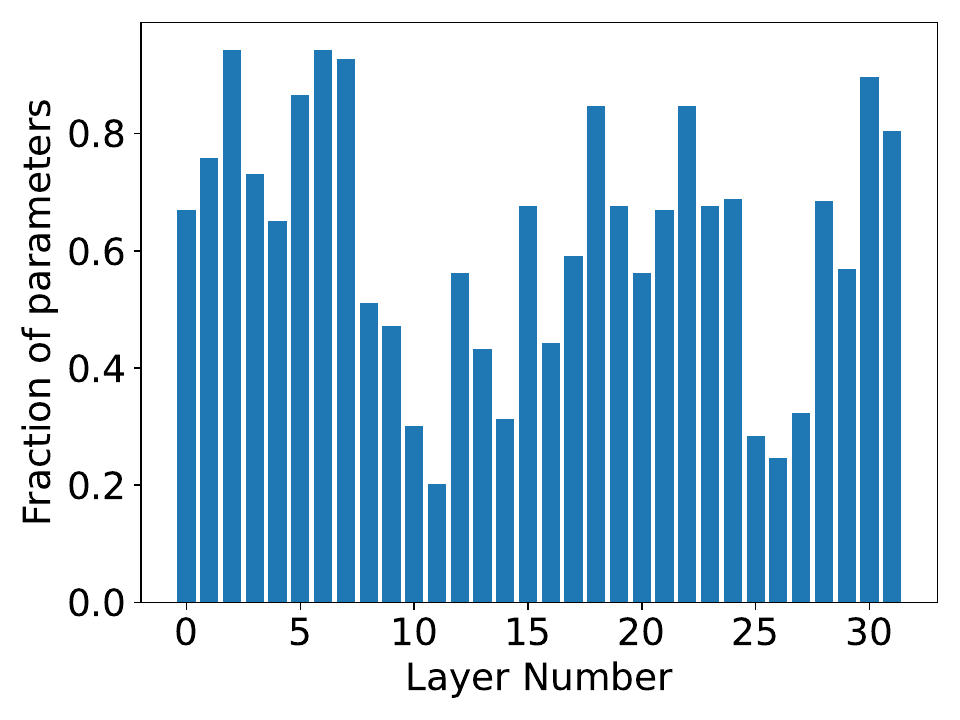}
       \caption{LLaMA-3.1-8B + GenRAG-I.}
        \label{fig:llama-8b-averages_by_layer-hotpot}
    \end{subfigure}
    \hfill
    \begin{subfigure}[b]{0.3\textwidth}
    \centering
    \includegraphics[width=1\textwidth]{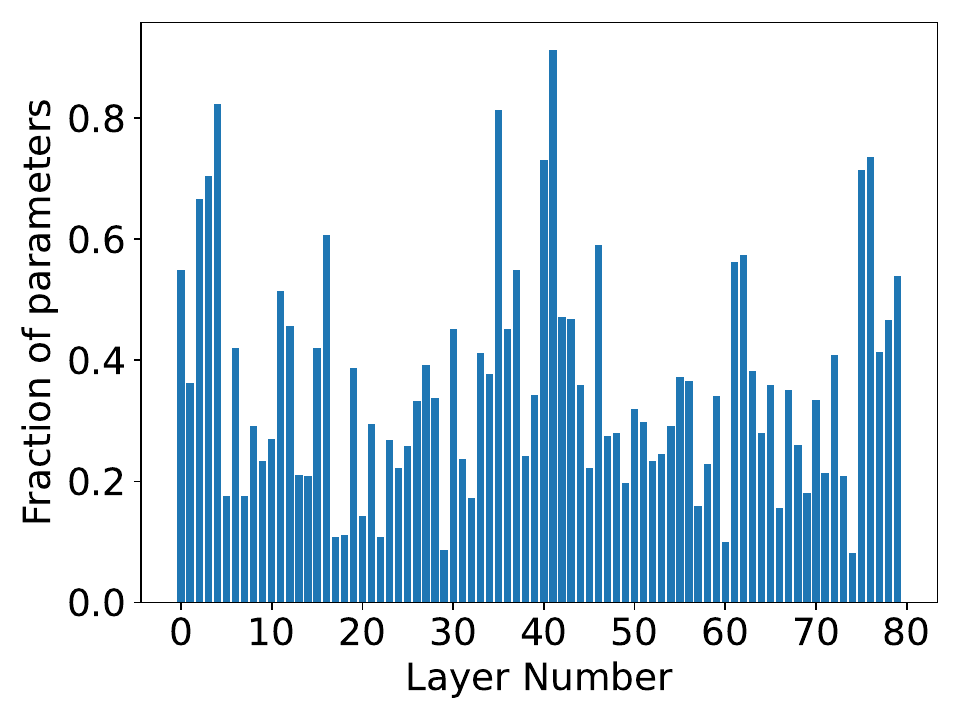}
        \caption{LLaMA-3.1-70B+GenRAG-I.}
        \label{fig:llama-70b-averages_by_layer-hotpot}
    \end{subfigure}
    \hfill
    \begin{subfigure}[b]{0.3\textwidth}
     \centering
    \includegraphics[width=1\textwidth]{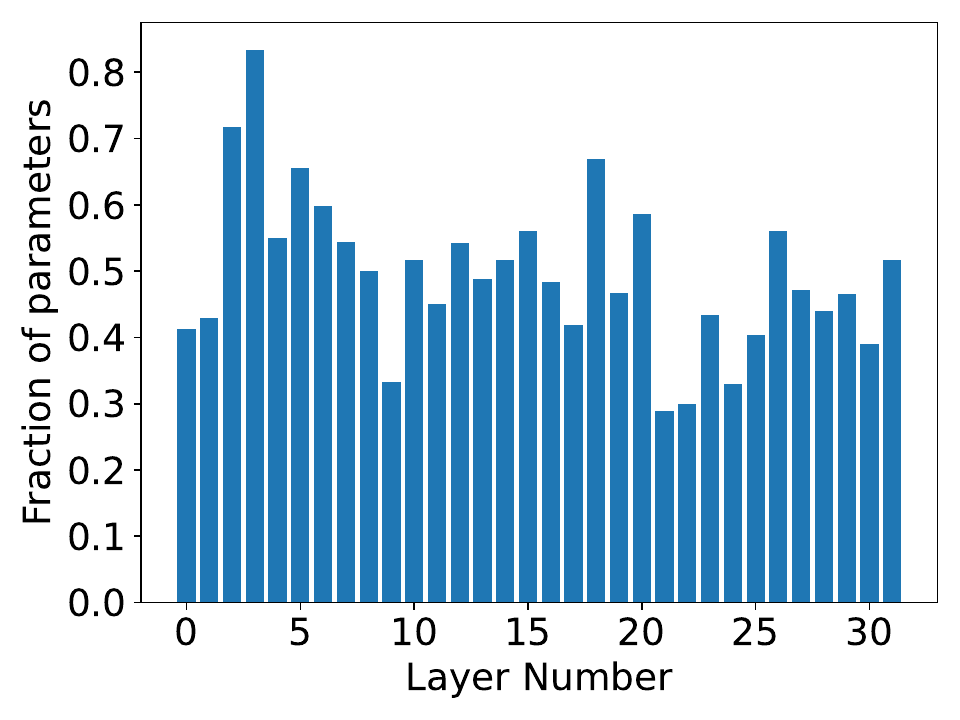}
       \caption{Phi-3-mini+MedRAG-I.}
        \label{fig:phi-mini-averages_by_layer-pubmed}
    \end{subfigure}
    \hfill
    \begin{subfigure}[b]{0.3\textwidth}
     \centering
    \includegraphics[width=1\textwidth]{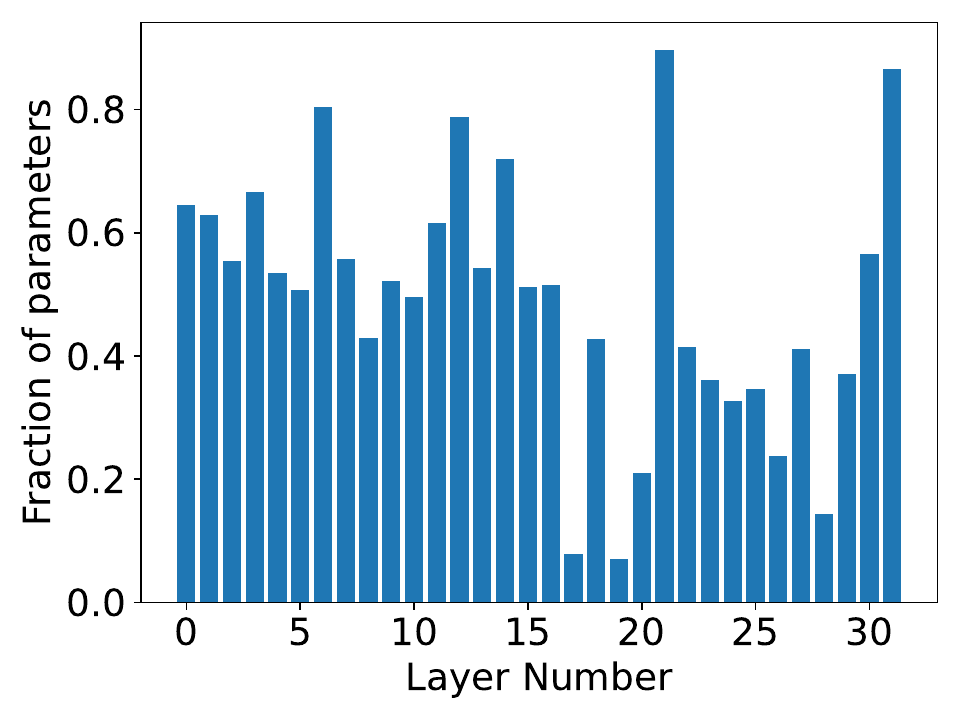}
       \caption{LLaMA-3.1-8B+MedRAG-I.}
        \label{fig:llama-8b-averages_by_layer-pubmed}
    \end{subfigure}
    \hfill
    \begin{subfigure}[b]{0.3\textwidth}
    \centering
    \includegraphics[width=1\textwidth]{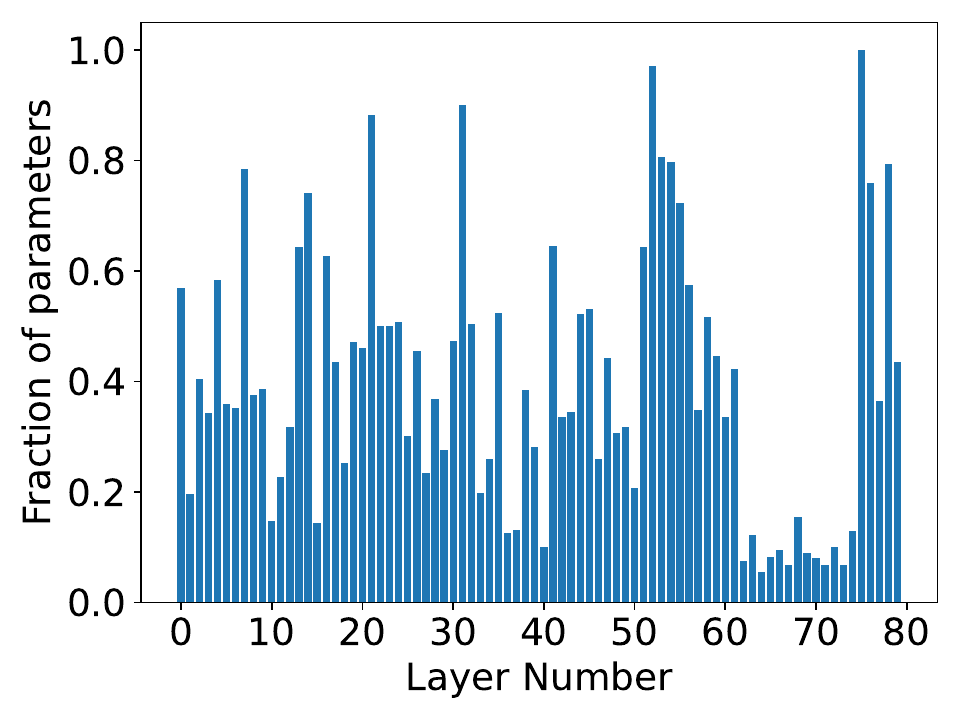}
        \caption{LLaMA-3.1-70B+MedRAG-I.}
        \label{fig:llama-70b-averages_by_layer-pubmed}
    \end{subfigure}
    \hfill
    \begin{subfigure}[b]{0.3\textwidth}
     \centering
    \includegraphics[width=1\textwidth]{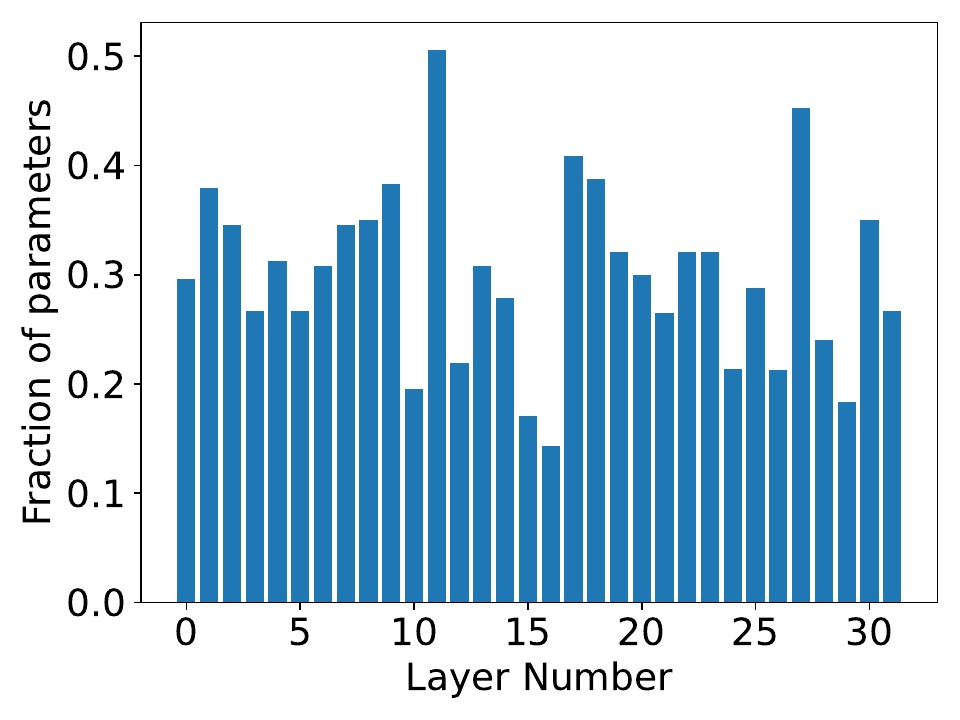}
       \caption{Phi-3-mini+Sentiment-I.}
        \label{fig:phi-mini-averages_by_layer-imdb}
    \end{subfigure}
    \hfill
    \begin{subfigure}[b]{0.3\textwidth}
     \centering
    \includegraphics[width=1\textwidth]{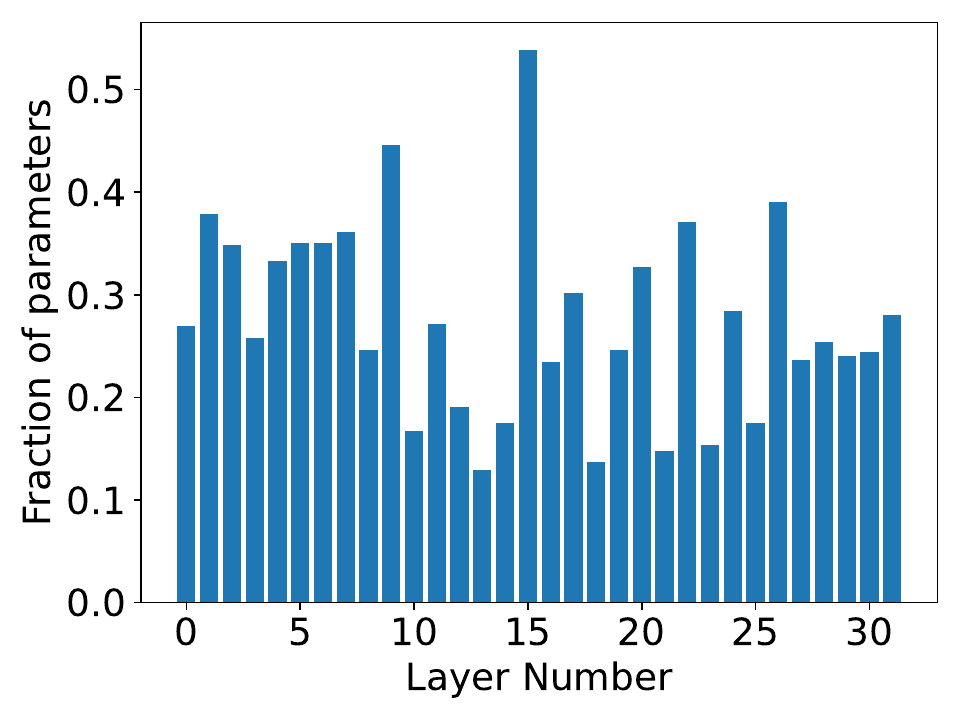}
       \caption{LLaMA-3.1-8B+Sentiment-I.}
        \label{fig:llama-8b-averages_by_layer-imdb}
    \end{subfigure}
    \hfill
    \begin{subfigure}[b]{0.3\textwidth}
    \centering
    \includegraphics[width=1\textwidth]{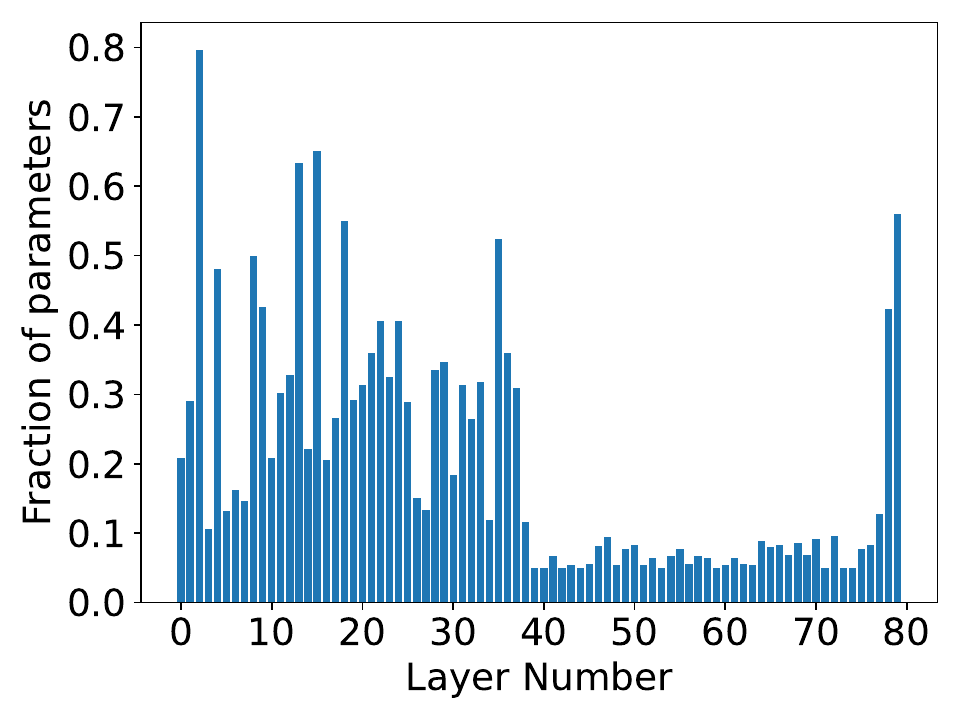}
        \caption{LLaMA-3.1-70B+Sentiment-I.}
        \label{fig:llama-70b-averages_by_layer-imdb}
    \end{subfigure}
    \caption{Fraction of remaining parameters for each layer in the model after conducing a Genetic Algorithm search (lower numbers imply higher compression).}
    \label{fig:appendix_genetic_algorithm_by_layer}
\end{figure}

Figure~\ref{fig:appendix_genetic_algorithm_by_layer} presents the average fraction of parameters retained per layer. For larger models such as LLaMA-3.1-70B, we observe a trend where the second half of layers tends to be significantly more compressible than the first half, with the exception of the final few layers. This pattern is especially pronounced in narrower, more specialized tasks like sentiment analysis (Fig. \ref{fig:llama-70b-averages_by_layer-imdb}). In contrast, for smaller models such as Phi-3-mini and LLaMA-3.1-8B, no consistent compressibility trend emerges across the different layers.

\clearpage

\subsection{Presence of Bottleneck Matrices.}
\label{appendix:bottleneck}
We perform a cross-population analysis of the top-performing individuals—defined as those within 20\% of the best recorded fitness—during our genetic algorithm search. For each matrix in every layer, we compute the probability that it remains unpruned (i.e., receives zero compression) across these individuals. As seen in Fig. ~\ref{fig:appendix_heatmap}, for LLaMA-3.1-70B, we consistently find matrices that are never pruned across all top individuals (highlighted in \textcolor{red}{red}), indicating that pruning these matrices leads to significant performance degradation. We refer to these as \textit{bottleneck matrices}. This phenomenon suggests that larger models tend to hyperspecialize certain matrices for critical sub-tasks and exhibit limited tolerance to pruning in those areas. We also find that these matrices fall into two categories: \textit{task-agnostic} (e.g., $w_{qkv}$ at $layer_{0}$ in LLaMA-3.1-70B), which resist pruning across all tasks and may encode general-purpose features such as punctuation handling, and \textit{task-specific} (e.g., $w_{1}$ at $layer_{3}$ in LLaMA-3.1-70B), which are indispensable only for particular tasks.

\begin{figure}[h!]
    \centering
    \begin{subfigure}[b]{0.45\textwidth}
        \centering
        \includegraphics[width=\textwidth]{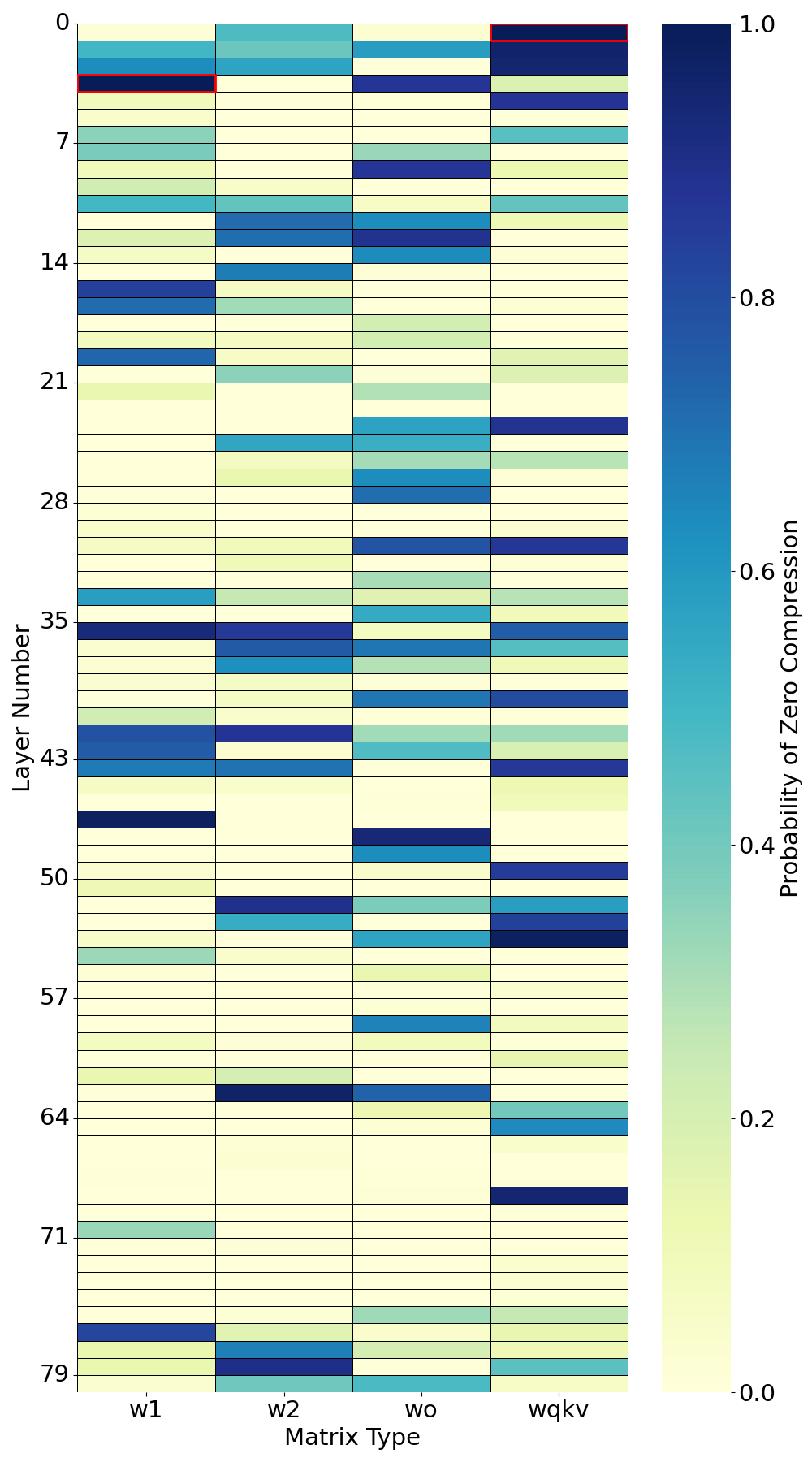}
        \caption{LLaMA-3.1-70B+GenRAG-I.}
        \label{fig:llama-70b-heatmap-hotpot}
    \end{subfigure}
    \hfill
    \begin{subfigure}[b]{0.45\textwidth}
        \centering
        \begin{subfigure}[b]{\textwidth}
            \centering
            \includegraphics[width=\textwidth]{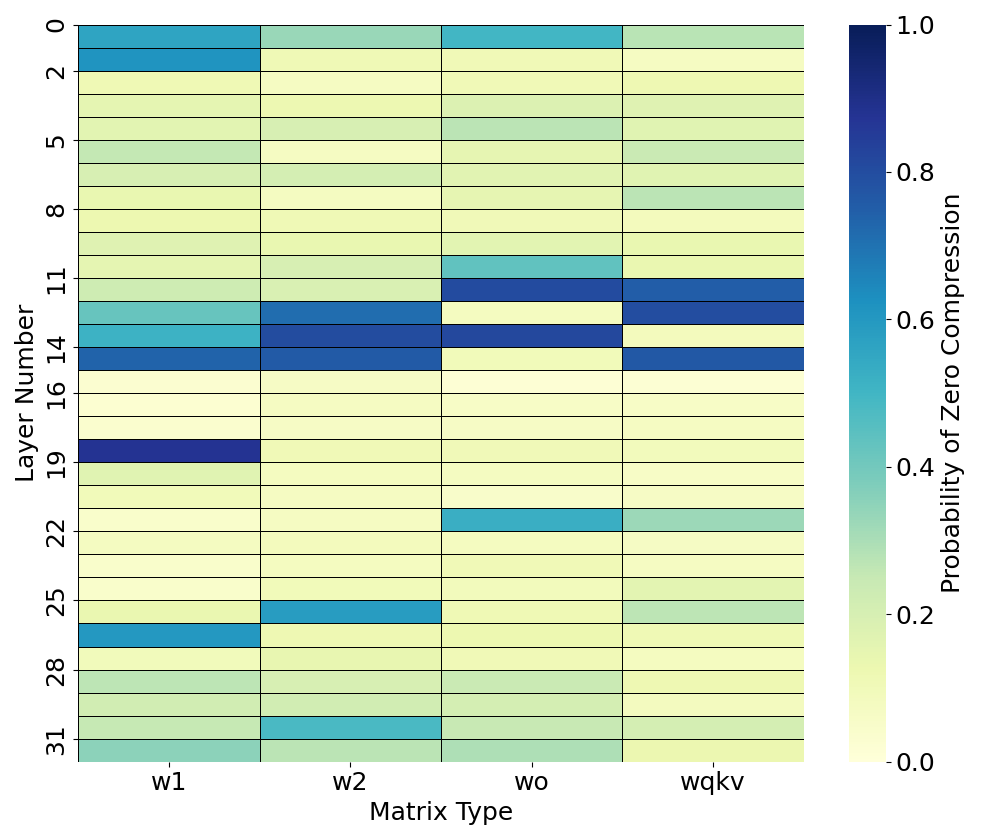}
            \caption{Phi-3-mini+GenRAG-I.}
            \label{fig:phi-mini-heatmap-hotpot}
        \end{subfigure}
        \vskip\baselineskip
        \begin{subfigure}[b]{\textwidth}
            \centering
            \includegraphics[width=\textwidth]{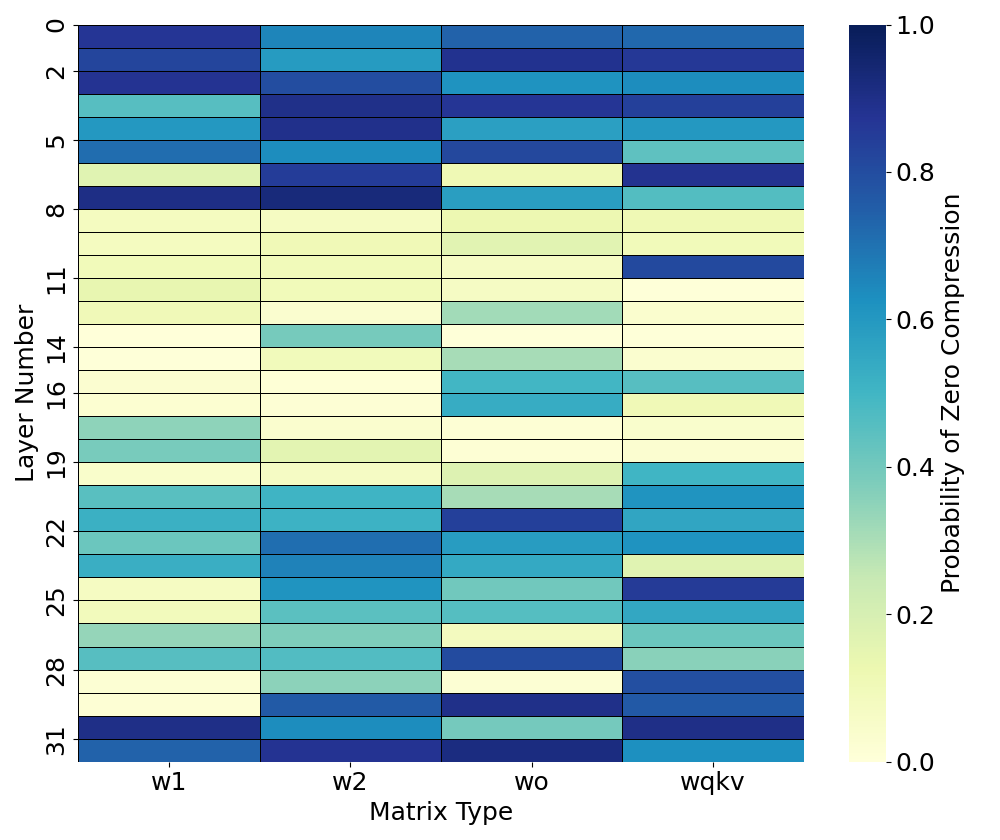}
            \caption{LLaMA-3.1-8B+GenRAG-I.}
            \label{fig:llama-8b-heatmap-hotpot}
        \end{subfigure}
    \end{subfigure}
    \caption{Heat map showing the bottleneck matrices during GA for different models/datasets. Boxes highlighted in \textcolor{red}{red} indicate that the matrix is unprunable (\ie, pruning results in significant loss).}
    \label{fig:appendix_heatmap}
\end{figure}

\begin{figure}[h!]
    \ContinuedFloat
    \centering
    \begin{subfigure}[b]{0.40\textwidth}
        \centering
        \includegraphics[width=0.9\textwidth]{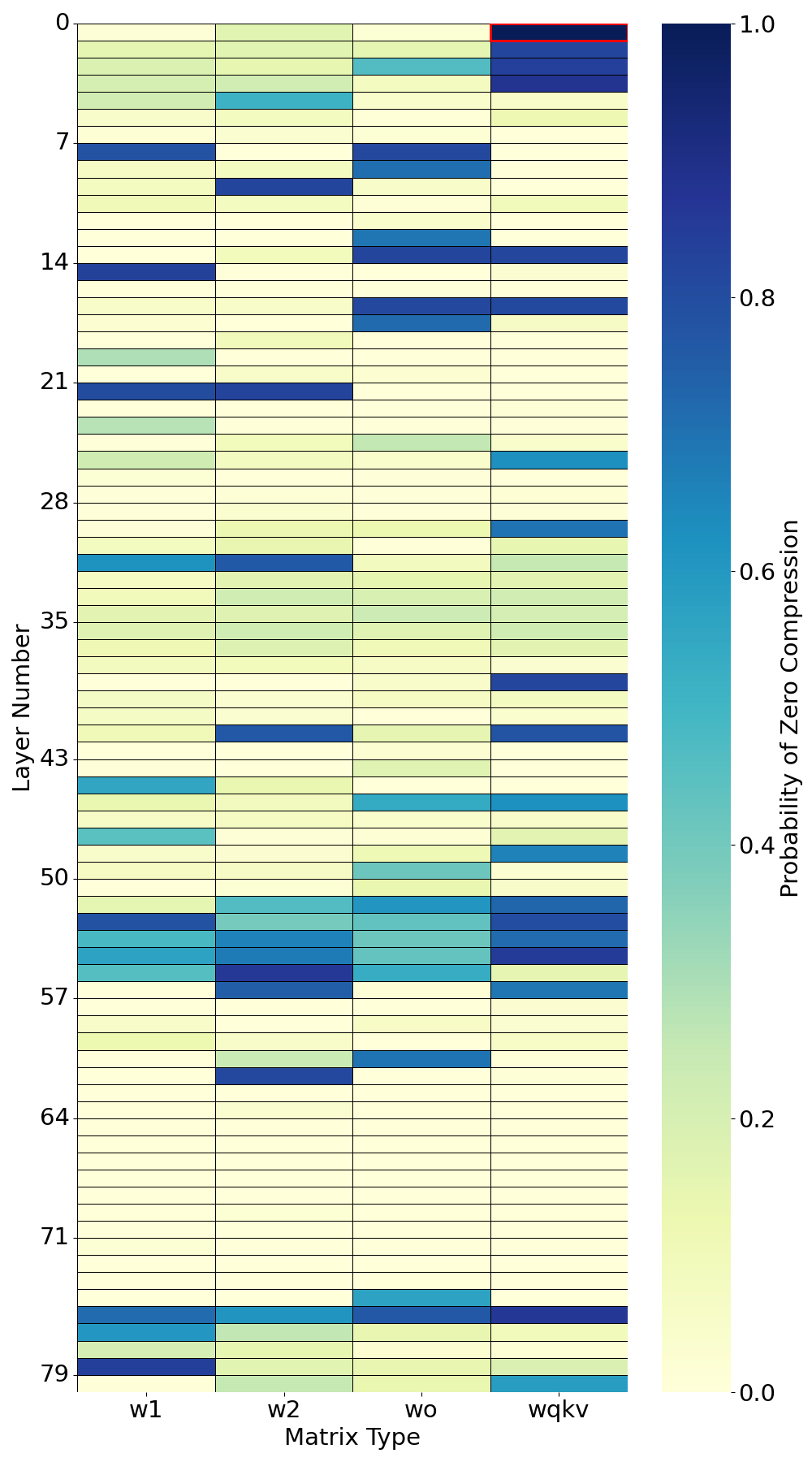}
        \caption{LLaMA-3.1-70B+MedRAG-I.}
        \label{fig:llama-70b-heatmap-pubmed}
    \end{subfigure}
    \hfill
    \begin{subfigure}[b]{0.40\textwidth}
        \centering
        \begin{subfigure}[b]{\textwidth}
            \centering
            \includegraphics[width=0.9\textwidth]{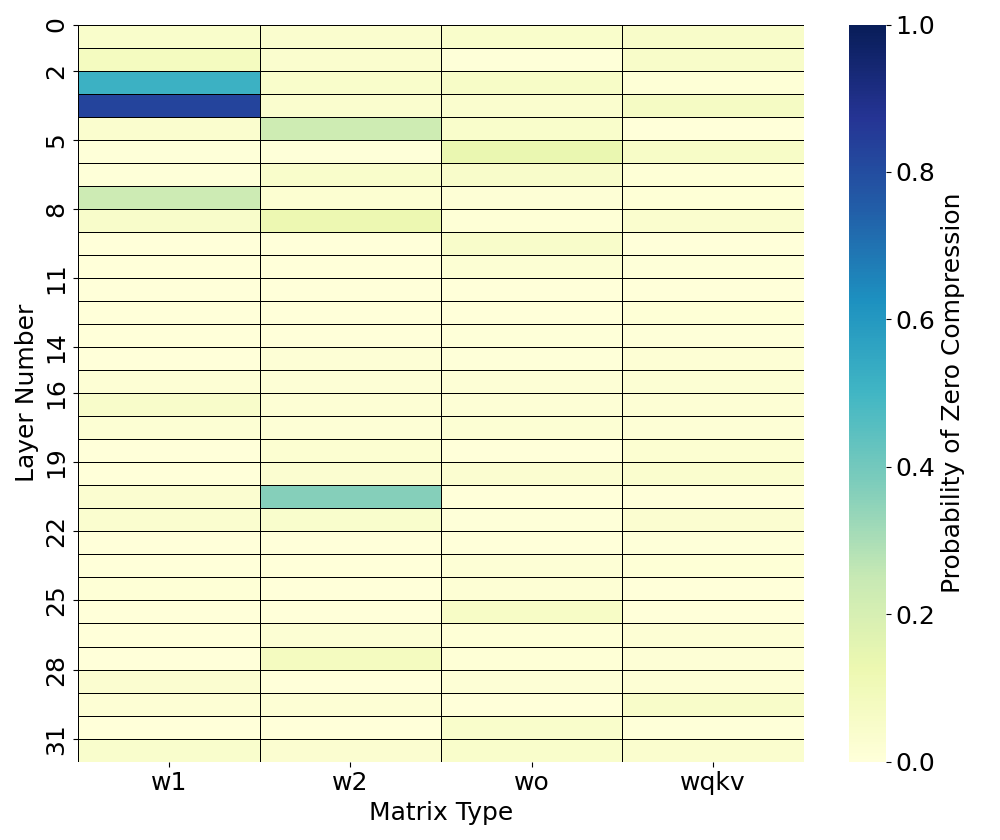}
            \caption{Phi-3-mini+MedRAG-I.}
            \label{fig:phi-mini-heatmap-pubmed}
        \end{subfigure}
        \vspace{0.5em}
        \begin{subfigure}[b]{\textwidth}
            \centering
            \includegraphics[width=0.9\textwidth]{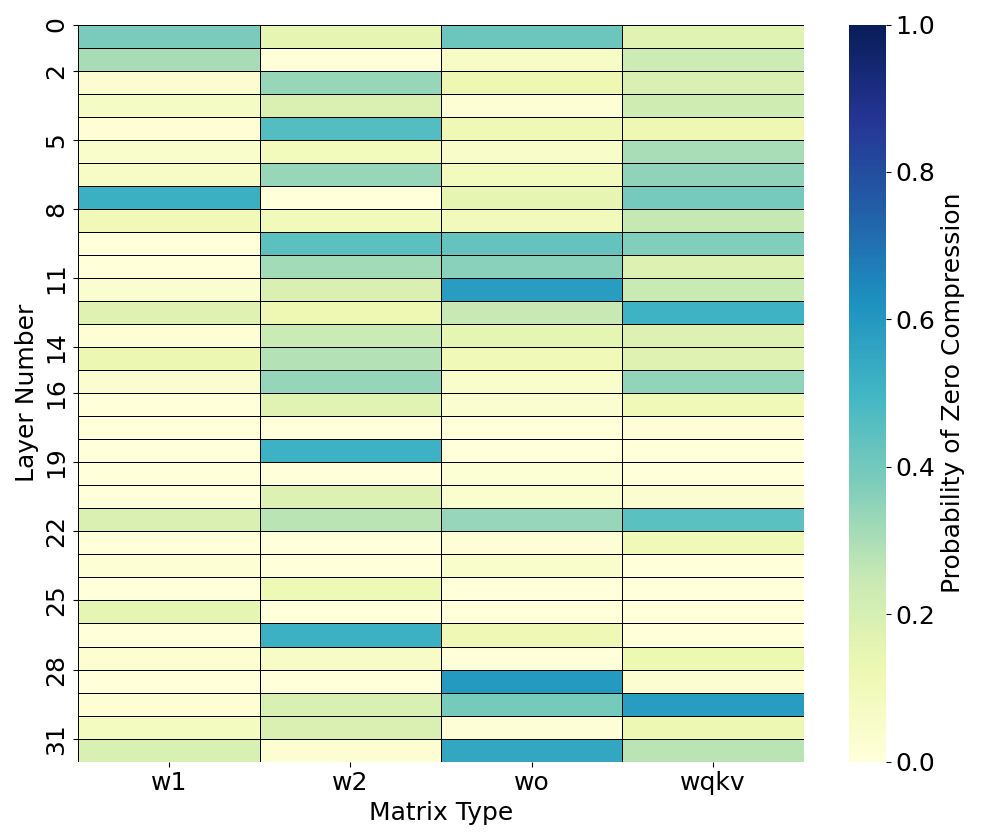}
            \caption{LLaMA-3.1-8B+MedRAG-I.}
            \label{fig:llama-8b-heatmap-pubmed}
        \end{subfigure}
    \end{subfigure}

    \vspace{1em} %

    \begin{subfigure}[b]{0.40\textwidth}
        \centering
        \includegraphics[width=0.9\textwidth]{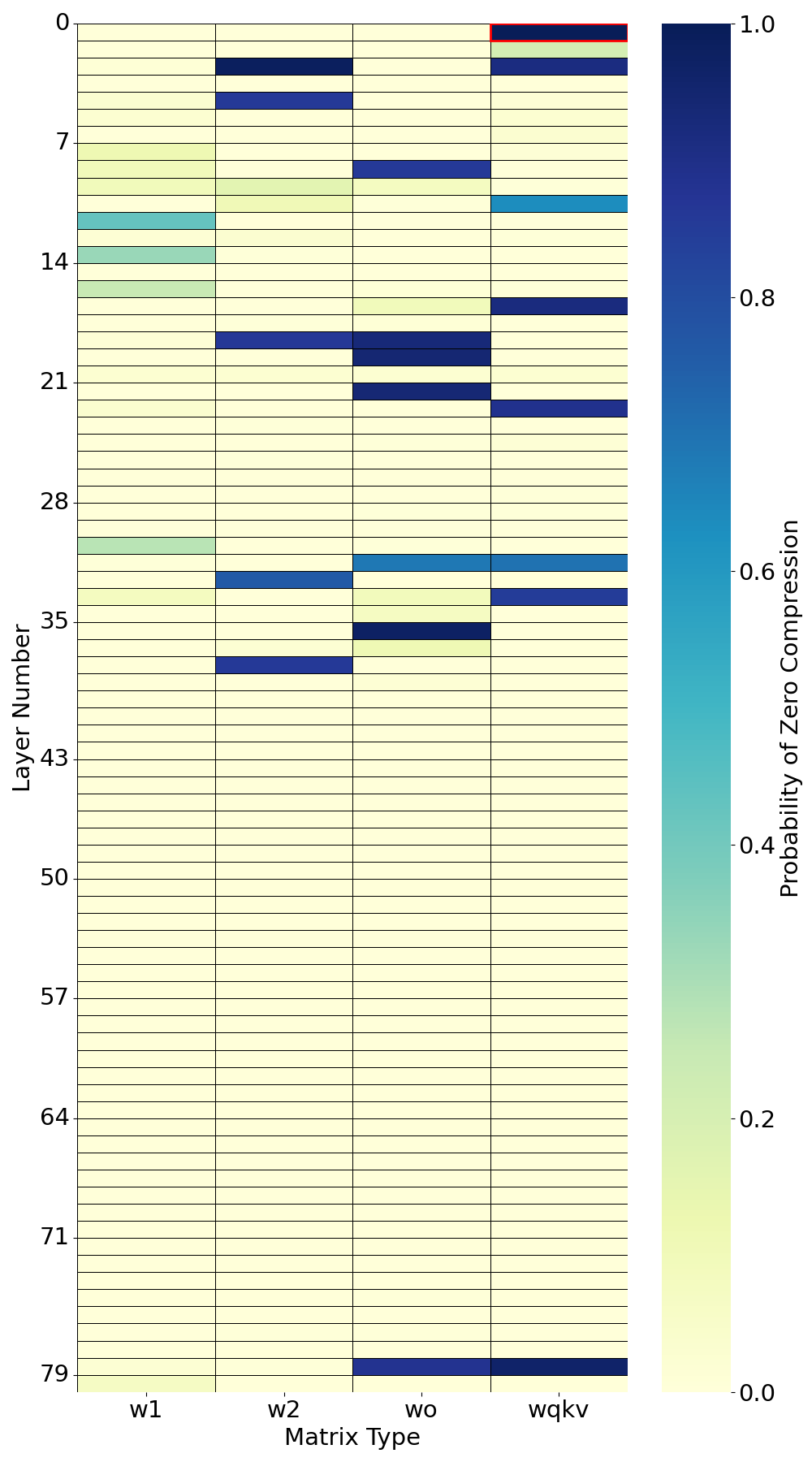}
        \caption{LLaMA-3.1-70B+Sentiment-I.}
        \label{fig:llama-70b-heatmap-imdb}
    \end{subfigure}
    \hfill
    \begin{subfigure}[b]{0.40\textwidth}
        \centering
        \begin{subfigure}[b]{\textwidth}
            \centering
            \includegraphics[width=0.9\textwidth]{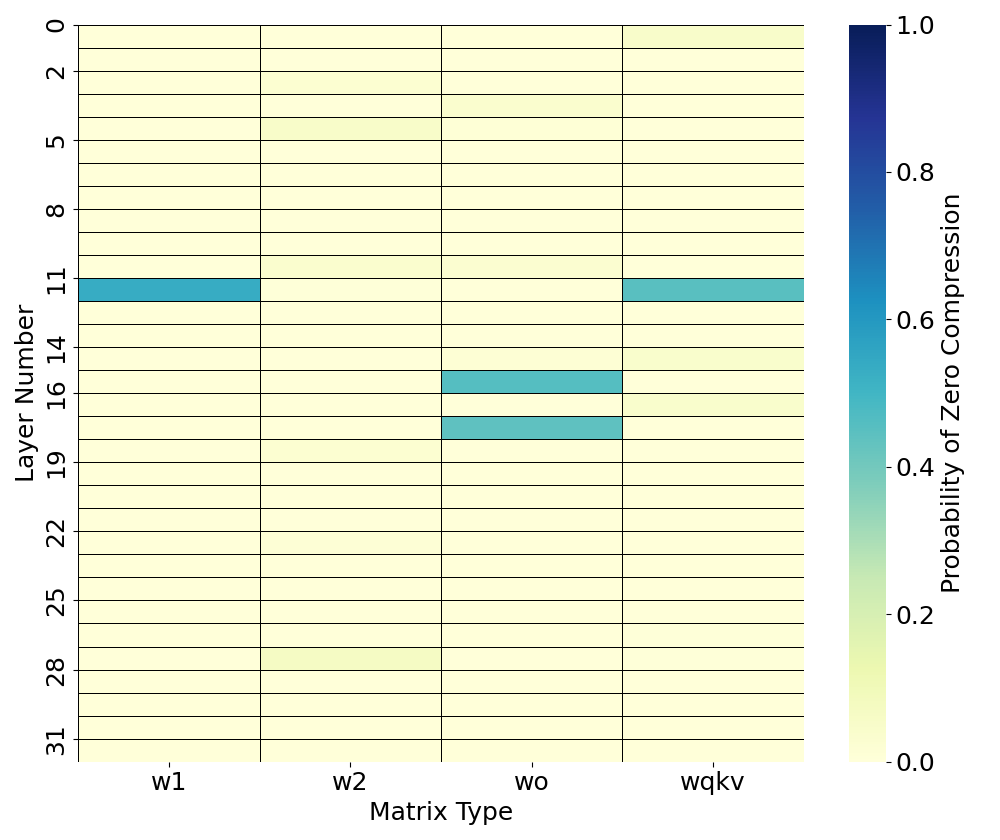}
            \caption{Phi-3-mini+Sentiment-I.}
            \label{fig:phi-mini-heatmap-imdb}
        \end{subfigure}
        \vspace{0.5em}
        \begin{subfigure}[b]{\textwidth}
            \centering
            \includegraphics[width=0.9\textwidth]{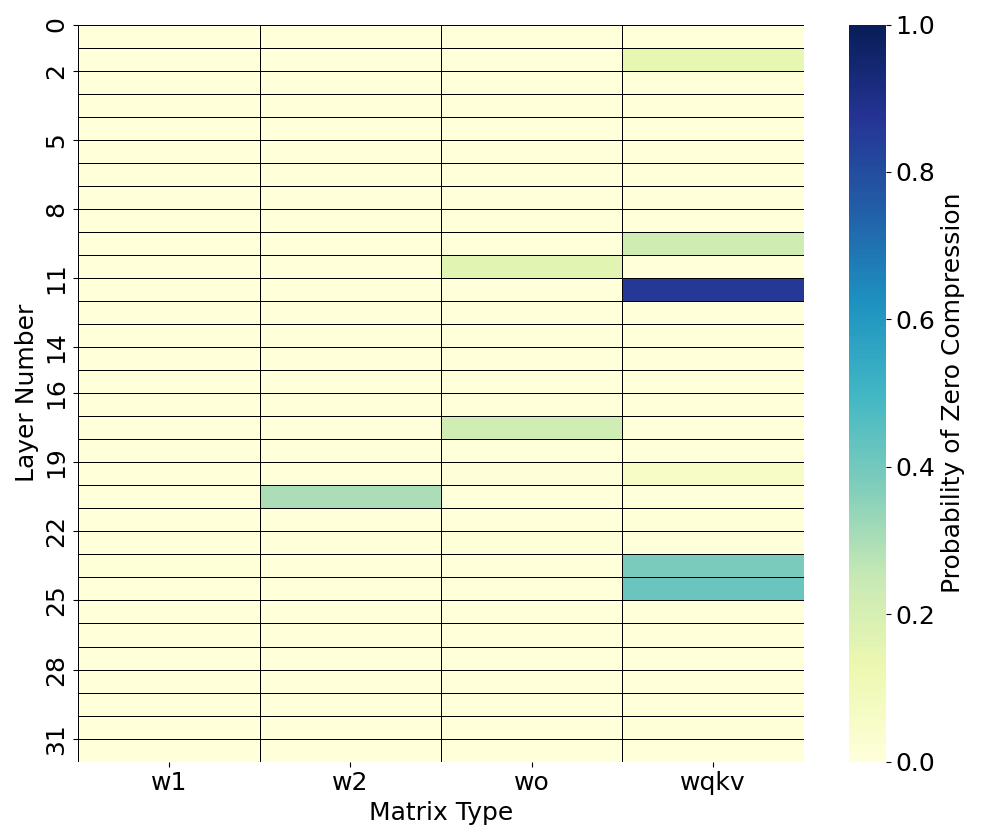}
            \caption{LLaMA-3.1-8B+Sentiment-I.}
            \label{fig:llama-8b-heatmap-imdb}
        \end{subfigure}
    \end{subfigure}

    \caption{(Continued)}
\end{figure}

\end{document}